\documentclass[11pt]{article}

\usepackage[preprint]{acl}

\usepackage{times}
\usepackage{latexsym}

\usepackage[T2A,T1]{fontenc}

\usepackage[utf8]{inputenc}

\usepackage{microtype}

\usepackage{inconsolata}

\usepackage{graphicx}

\usepackage{url}
\usepackage{booktabs}
\usepackage{CJKutf8}
\usepackage{multirow}
\usepackage[russian,english]{babel}
\usepackage{amsmath,amsfonts}

\newsavebox{\SWEETtablebox}
\newenvironment{SWEETfitwidth}
  {\begin{lrbox}{\SWEETtablebox}}
  {\end{lrbox}%
   \ifdim\wd\SWEETtablebox>\linewidth
     \resizebox{\linewidth}{!}{\usebox{\SWEETtablebox}}%
   \else
     \usebox{\SWEETtablebox}%
   \fi}

\title{SWEET: Mitigating Spurious Feature Reliance in LLM Reasoning via Weight Editing}

\author{
\textbf{Zhizhang Fu\textsuperscript{1,*},
Lucas Zhao\textsuperscript{2,*},
Michalis Korakakis\textsuperscript{3,4},
Mengru Ding\textsuperscript{5},} \\
\textbf{Lukasz Szpruch\textsuperscript{4,6},
Yue Zhang\textsuperscript{1,\textdagger}} \\
\textsuperscript{1}Westlake University,
\textsuperscript{2}Yale University,
\textsuperscript{3}University of Cambridge, \\
\textsuperscript{4}The Alan Turing Institute,
\textsuperscript{5}Xianyu of Alibaba,
\textsuperscript{6}University of Edinburgh \\
{\small \textsuperscript{*}Equal contribution.
\textsuperscript{\textdagger}Corresponding author: \texttt{zhangyue@westlake.edu.cn}.}
}

\begin{document}

\maketitle

\begin{abstract}
Reasoning in large language models can be disrupted by simple input perturbations (e.g.\ adding irrelevant distractors), suggesting reliance on spurious correlations. Identifying and suppressing the spurious features involved in long chain-of-thought reasoning remains challenging. We introduce Spurious-feature WEight EdiTing~(SWEET), a training-free method that combines counterfactual interventions with mechanistic interpretability to identify and suppress spurious-features. SWEET extracts token- and sequence-level behavioral directions from activation contrasts between original problems and their counterfactual variants---capturing local and global patterns of spurious reasoning. It then localizes the responsible MLP neurons and attention heads and suppresses their contributions along these directions through weight edits. Using only 20 original--counterfactual pairs for direction extraction, SWEET improves accuracy by an average of 11.8 percentage points on perturbed inputs and 2 points on original inputs across five open-weight reasoning models and five reasoning benchmarks. It also reduces the relative gap between original and perturbed accuracies from 17.6\% to 2.5\%. Analyses identify two recurring forms of spurious reasoning, i.e.\ over-integrating irrelevant conditions and overthinking, and provide mechanistic evidence that SWEET improves reasoning robustness by suppressing both behavioral directions.
\end{abstract}

\begin{figure*}[t]
    \centering
    \includegraphics[width=0.82\textwidth]{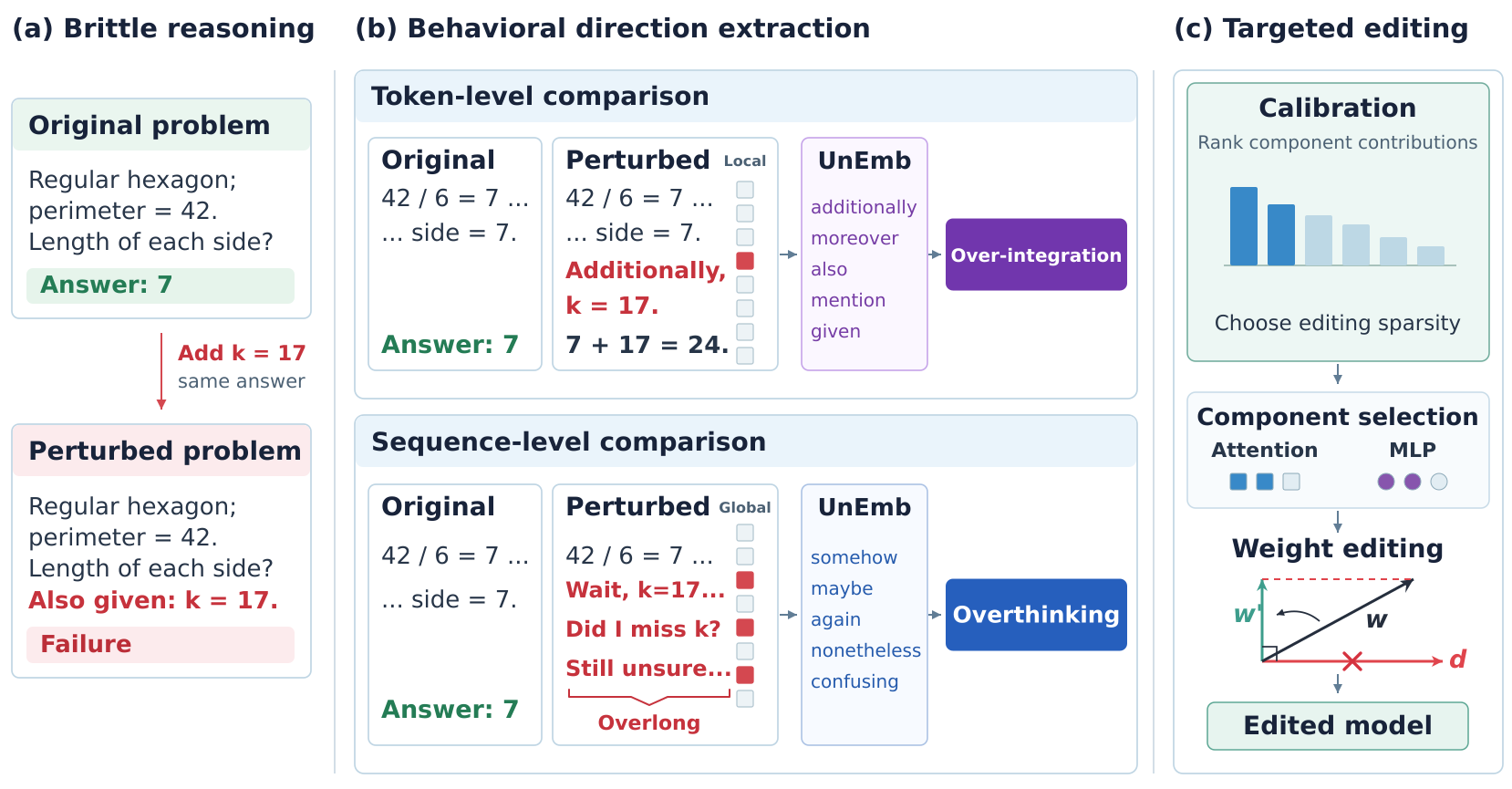}
    \caption{
    Overview of SWEET. (a)  SWEET constructs answer-preserving perturbations by adding irrelevant conditions, then selects original--perturbed pairs for which the model answers only the original problem correctly. (b) Token-level extraction contrasts activations along the same reasoning trace under both prompts, focusing on positions most affected by the perturbation to capture the incorporation of irrelevant information (\emph{over-integration}). Sequence-level extraction contrasts mean activations from independently generated traces, capturing broader behavioral changes, illustrated here by repeated doubt and an unnecessarily long response (\emph{overthinking}). Red highlights mark these behaviors; unembedding (UnEmb) word lists help interpret the extracted directions. (c) SWEET ranks attention heads and MLP neurons by their contributions to these directions, selects the number of components to edit using calibration data, and modifies their output weights to remove contributions along the extracted directions.
    }
    \label{fig:head}
\end{figure*}

\section{Introduction}
\label{intro}
Large language models~(LLMs) have made remarkable progress on reasoning  tasks~\citep{openai2025o3,guo2025deepseek,team2026kimi}. However, even strong models can fail under simple perturbations, such as rephrasing problem scenarios, changing numerical values or adding distractors~\citep{huang2026thinkbench,mirzadeh2025gsm}. This brittleness suggests reliance on spurious correlations~\citep{geirhos2020shortcut}, making performance sensitive to how problems are presented rather than solely to the reasoning required.  As LLMs are increasingly used for scientific problem solving and safety-critical decision-making, such failures can undermine scientific conclusions and lead to harmful decisions.

Prior work on mitigating spurious correlations has primarily focused on discriminative tasks, where models predict a single label~\citep{mahabadi2020end,wang2022identifying,zhou2025fighting}, rather than long chain-of-thought~(CoT) reasoning~\citep{chen2026towards}. In long CoT, an error caused by a spurious feature at one step can propagate through the remaining trace, obscuring the source of the final failure. Recent methods address spurious-feature reliance in CoT through token pruning and distillation or causal interventions~\citep{wu2024decot,xie2025mitigating,guo2026learning}. Although these methods improve accuracy, such gains alone do not establish which forms of spurious-feature reliance, if any, have been mitigated. Without identifying the behaviors and internal components responsible for spurious reasoning, it remains unclear whether interventions correct the underlying failure mechanisms or compensate for their effects~\citep{mendelson-belinkov-2021-debiasing}.

To bridge this gap, our central intuition is that identifying and characterizing spurious features not only provides an interpretable account of reasoning failures, but also enables us to directly suppress their influence, thereby increasing the contribution of causally relevant features and improving reasoning robustness. We introduce Spurious-feature WEight EdiTing~(SWEET), a training-free method for identifying and suppressing spurious-feature reliance in LLM reasoning. Motivated by Pearl's causal hierarchy~\citep{pearl2009causality}, we treat answer-preserving perturbations as counterfactual interventions on task-irrelevant aspects of the input (Figure~\ref{fig:head}(a)). We contrast paired reasoning activations at token and sequence levels to extract behavioral directions associated with local and global patterns of spurious reasoning, respectively (Figure~\ref{fig:head}(b)). We then localize responsible attention heads and MLP neurons and remove their contributions along these directions through weight edits (Figure~\ref{fig:head}(c)). The edited model can be applied to subsequent problems without further modification or training, with improved robustness to spurious features. To the best of our knowledge, SWEET is the first method to combine counterfactual interventions with mechanistic interpretability to identify and suppress spurious-feature reliance at both token and sequence levels in long-CoT reasoning.

We evaluate SWEET on five open-weight reasoning models across five reasoning benchmarks and their distractor-augmented variants. Using up to 20 original--perturbed pairs to extract behavioral directions, SWEET improves accuracy by an average of 11.8 percentage points on perturbed inputs and 2 points on original inputs, while reducing the accuracy gap between original and perturbed inputs from 17.6\% to 2.5\%. These results demonstrate SWEET's effectiveness across model families, scales, and reasoning benchmarks.

We further analyze the extracted behavioral directions to characterize spurious reasoning and understand SWEET's robustness gains. Across five models, we identify two recurring patterns: \emph{over-integration}, in which models incorporate solution-relevant conditions and irrelevant distractors into their reasoning without distinguishing between them, and \emph{overthinking}, in which models repeatedly question, revise, or abandon otherwise viable reasoning steps. Causal interventions show that editing the corresponding directions suppresses these behaviors, providing mechanistic evidence that SWEET improves robustness by reducing these two forms of spurious reasoning.

We release our code at
\href{https://anonymous.4open.science/r/SpuriousFeatureWeightEditing-5C29/}{this anonymous repository}.

\section{Preliminaries}
\label{sec:preliminaries}

\paragraph{Steering directions.}
\label{sec:steering_directions}

Behavioral attributes of language models can be associated with
directions in hidden-state
space~\citep{turner2023steering,rimsky2024steering,chen2025persona},
commonly estimated by contrasting mean activations between behaviors.
For example, contrasting harmful requests that models refuse with
harmless requests they answer yields a refusal
direction~\citep{arditi2024refusal}, while contrasting
``Think Step by Step'' and ``Answer Immediately'' prompts yields a
CoT steering vector~\citep{zhang2024uncovering}.
Let $h_\ell(s)$ denote the residual-stream activation at layer $\ell$
at the final token position of context $s$.
Activation steering modifies this representation as
$h_\ell'(s)=h_\ell(s)+\alpha v_\ell$,
where $v_\ell$ is a steering direction extracted by contrasting
mean activations, and $\alpha$ controls the intervention's
sign and magnitude.

\paragraph{Residual-stream decomposition.}
\label{sec:prelim_units}
Attention heads and MLP neurons contribute additively to the
residual stream~\citep{elhage2021mathematical,geva2021transformer}.
We refer to each head or neuron as a \emph{unit}.
For an input context $s$, their contributions at the current token
position are
\begin{equation}
\begin{gathered}
c_{\ell,h}^{\mathrm{Attn}}(s)
= W_O^{\ell,h}z_{\ell,h}(s),
\\
c_{\ell,j}^{\mathrm{MLP}}(s)
= a_{\ell,j}(s)W_{\mathrm{down}}^\ell[:,j].
\end{gathered}
\label{eq:prelim_unit_outputs}
\end{equation}
Here, $z_{\ell,h}(s)$ is the output of attention head $h$ before
the output projection, and $W_O^{\ell,h}$ is the
column block of the attention output matrix $W_O^\ell$.
For MLP neuron $j$, $a_{\ell,j}(s)$ is its scalar activation
entering the output projection, including the gating operation
in gated MLPs, and $W_{\mathrm{down}}^\ell[:,j]$ is its output
weight vector.
We use $c_u(s)\in\mathbb R^{d_{\mathrm{model}}}$ to denote
either type of unit contribution.

\paragraph{Directional attribution.}
\label{sec:prelim_attribution}
Direct logit attribution (DLA) measures a component's direct
contribution to a target token's logit by projecting its
residual-stream output toward the token's unembedding
direction~\citep{elhage2021mathematical}.
Following this projection principle, a unit's contribution along
any normalized direction $d_\ell^g$ in the same layer's residual
space can be expressed as~\citep{sun2025thinkedit}
\begin{equation}
A_u^g(s)=(d_\ell^g)^\top c_u(s),
\label{eq:prelim_direction_attribution}
\end{equation}
where $u$ belongs to layer $\ell$ and $g$ indexes the direction.
Positive scores indicate contributions along the direction;
negative scores indicate contributions against it.

\section{Spurious-feature Weight Editing}
\label{sec:method}

SWEET (1) extracts behavioral directions associated with spurious-feature reliance, (2) identifies responsible MLP neurons and attention heads, and (3) suppresses their contributions through weight edits to improve reasoning robustness. We describe original--perturbed pair construction and selection (Section~\ref{sec:pair_selection}), behavioral direction extraction (Section~\ref{sec:direction_extraction}), component ranking and selection (Section~\ref{sec:component_ranking}), directional weight editing (Section~\ref{sec:weight_editing}), and component assignment and sparsity calibration (Section~\ref{sec:sparsity_calibration}).

\subsection{Original--Perturbed Pairs}
\label{sec:pair_selection}
For each LLM to be edited with SWEET, we construct original--perturbed problem pairs using NoOp~\citep{mirzadeh2025gsm}, which adds irrelevant information to an original problem $x^{\mathrm o}$ to obtain $x^{\mathrm p}$ without changing the correct answer. We use that LLM to generate responses to both versions and retain pairs for which it answers the original problem correctly but the perturbed problem incorrectly. We sample up to $N_{\mathrm{ext}}$ qualifying pairs (or all if fewer qualify) to form $\mathcal D_{\mathrm{ext}}=\{(x_i^{\mathrm o},x_i^{\mathrm p})\}_{i=1}^{N}$. Extraction and calibration data are excluded from final evaluation.

\subsection{Steering Direction Extraction}
\label{sec:direction_extraction}
To characterize spurious reliance at both local and global levels,
we extract two complementary behavioral directions: a token-level
(TL) direction representing localized perturbation-related features
and a sequence-level (SL) direction representing broader
trajectory-level features.
To improve the stability of direction estimation, we normalize activation differences by their across-pair variability.

Let $y_i^{\mathrm o}$ and $y_i^{\mathrm p}$ denote the model's
responses to the original and perturbed problems, respectively.
For per-pair activation difference vectors
$\{\delta_{\ell,i}\}_{i=1}^{N}$ at layer $\ell$, let $\mu_\ell$
and $\sigma_\ell$ be their dimension-wise mean and standard
 deviation over the $N$ pairs, and define
\begin{equation}
\mathcal V_\ell\bigl[\{\delta_{\ell,i}\}_{i=1}^{N}\bigr]
=\mu_\ell\oslash\bigl(\sigma_\ell+
\lambda\operatorname{med}(\sigma_\ell)\bigr),
\label{eq:variance_normalisation}
\end{equation}
where $\oslash$ denotes dimension-wise division and
$\operatorname{med}(\sigma_\ell)$ is the median entry of
$\sigma_\ell$.
This scaling downweights dimensions with high across-pair
variability, while $\lambda$ controls the strength of the
stabilizing term to limit amplification when a dimension's
standard deviation is near zero.

\paragraph{Token-level extraction.}
We teacher-force each reference
trace $y_i^r$ under both prompts and compute the next-token divergence
$D_{i,t}^{r}=D_{\mathrm{KL}}\big(p_\theta(\cdot\mid x_i^{\mathrm p},y_{i,<t}^{r})\,\|\,p_\theta(\cdot\mid x_i^{\mathrm o},y_{i,<t}^{r})\big)$.
We select the top-ranked positions by divergence to form $S_i^r$
and average the activation differences taken immediately before
predicting $y_{i,t}^r$ over those positions:
\begin{equation}
\begin{aligned}
\delta_{\ell,i}^{r}
&=\frac{1}{|S_i^r|}\sum_{t\in S_i^r}
\left[h_\ell(x_i^{\mathrm p},y_{i,<t}^{r})-h_\ell(x_i^{\mathrm o},y_{i,<t}^{r})\right].
\end{aligned}
\label{eq:token_delta}
\end{equation}
The two reference traces can yield
different activation-difference distributions across pairs.
We therefore normalize the activation differences for each
reference type separately before averaging the resulting directions:
\begin{equation}
v_\ell^{\mathrm{TL}}
=\frac{1}{2}\sum_{r\in\{\mathrm o,\mathrm p\}}
\mathcal V_\ell\bigl[\{\delta_{\ell,i}^{r}\}_{i=1}^{N}\bigr].
\label{eq:token_direction}
\end{equation}
Fixing the prefix separates prompt-induced changes from differences in generated
history, and using both reference traces reduces dependence on either. Each
nonzero layerwise vector is then normalized, $d_\ell^g=v_\ell^g/\|v_\ell^g\|_2$
for $g\in\{\mathrm{SL},\mathrm{TL}\}$.

\paragraph{Sequence-level extraction.}
Following contrastive CoT steering~\citep{zhang2024uncovering}, we mean-pool
the residual-stream activation $h_\ell(x_i^r,y_{i,\le t}^r)$ after each of the
$T_i^r$ response tokens (excluding the prompt) of each independently generated
trace, $r\in\{\mathrm o,\mathrm p\}$, and apply
Equation~\ref{eq:variance_normalisation} to the per-pair differences:
\begin{equation}
\begin{gathered}
v_\ell^{\mathrm{SL}}
=\mathcal V_\ell\bigl[\{\bar h_{\ell,i}^{\mathrm p}-\bar h_{\ell,i}^{\mathrm o}\}_{i=1}^{N}\bigr],
\\
\bar h_{\ell,i}^{r}=\frac{1}{T_i^{r}}\sum_{t=1}^{T_i^{r}}h_\ell(x_i^{r},y_{i,\le t}^{r}).
\end{gathered}
\label{eq:sequence_direction}
\end{equation}
This captures broad trajectory differences without requiring token alignment.

\subsection{Component Ranking and Selection}
\label{sec:component_ranking}
For each direction, we score every unit using
Equation~\ref{eq:prelim_direction_attribution}.
We first average over the token contexts $\mathcal C_q^g$
within each attribution trace, then over the $M_g$ traces
from the extraction set, giving each trace equal weight:
\begin{equation}
\bar A_u^g =
\frac{1}{M_g}\sum_{q=1}^{M_g}
\frac{1}{|\mathcal C_q^g|}
\sum_{s\in\mathcal C_q^g} A_u^g(s).
\label{eq:mean_direction_attribution}
\end{equation}
For $g\in\{\mathrm{SL},\mathrm{TL}\}$ and unit type
$m\in\{\mathrm{MLP},\mathrm{Attn}\}$, let $\mathcal U_{g,m}^{+}$ contain the
units of type $m$, across all layers, with $\bar A_u^g>0$. Each of the four
direction--type groups has its own budget: we select the top fraction $\tau$,
\begin{equation}
\begin{gathered}
\mathcal E_{g,m}(\tau)=\operatorname{Top}_{k}\left(\mathcal U_{g,m}^{+};\bar A^g\right),
\\ k=\max\bigl(1,\lfloor\tau|\mathcal U_{g,m}^{+}|\rfloor\bigr)\text{ for }\tau>0.
\end{gathered}
\label{eq:selected_components}
\end{equation}
Heads are ranked only against heads and neurons against neurons; pooling them
would let $\tau$ shift the edit between unit types rather than vary its extent.

\subsection{Directional Weight Editing}
\label{sec:weight_editing}
We adapt weight orthogonalization~\citep{arditi2024refusal} to the selected
units. With $W_u=W_{\mathrm{down}}^\ell[:,j]$ for
neuron $j$ or $W_O^{\ell,h}$ for head $h$,
\begin{equation}
W_u'=\left(I-d_\ell^g(d_\ell^g)^\top\right)W_u .
\label{eq:weight_edit}
\end{equation}
For any input $a_u$ (a neuron activation or a head's pre-projection output), the
edited output $c_u'=W_u'a_u$ satisfies $(d_\ell^g)^\top c_u'=0$, while its
component orthogonal to $d_\ell^g$ is unchanged. Unselected units are untouched,
and the edit is applied once to the weights, requiring neither training nor
activation interventions at inference.

\subsection{Component Assignment and Sparsity Calibration}
\label{sec:sparsity_calibration}
On a calibration set $\mathcal D_{\mathrm{cal}}$ disjoint from the test set, with directions and rankings fixed, we sweep a grid
$\mathcal T_{g,m}$ of sparsities for each group in isolation, editing only
$\mathcal E_{g,m}(\tau)$. Each candidate is scored by
$\mathcal S_{g,m}(\tau)=\frac12[\operatorname{Acc}_{\mathrm o}+\operatorname{Acc}_{\mathrm p}]$
on $\mathcal D_{\mathrm{cal}}$, so a sparsity that narrows the gap by degrading
original-problem accuracy is not rewarded. To limit model-specific tuning on
the calibration data, we choose one $\tau$ per group by charging each candidate
the accuracy it gives up relative to each model's own best:
\begin{equation}
\begin{aligned}
R_{g,m}(\tau)&=\frac{1}{|\mathcal K|}\sum_{k\in\mathcal K}
\Bigl(\max_{\tau'\in\mathcal T_{g,m}}\mathcal S^{(k)}_{g,m}(\tau')-\mathcal S^{(k)}_{g,m}(\tau)\Bigr),
\\
\tau^*_{g,m}&=\arg\min_{\tau\in\mathcal T_{g,m}}R_{g,m}(\tau),
\end{aligned}
\label{eq:regret}
\end{equation}
where $\mathcal K$ is the set of models. Regret is zero only for a $\tau$ optimal
on every model; otherwise it is the cost, in accuracy points, of one global choice.

We apply the SL edit to attention heads and the TL edit to MLP neurons.
This assignment aligns SL's trajectory-level contrasts with attention's
cross-position information aggregation, and TL's position-level contrasts
with the position-wise transformations performed by
MLPs~\citep{vaswani2017attention}.
Appendix~\ref{app:tl-sl-allocation}
(Table~\ref{tab:tl-sl-allocation}) compares this assignment with applying
both directions to both unit types.
The calibrated sparsities
$(\tau^*_{\mathrm{SL,Attn}},\tau^*_{\mathrm{TL,MLP}})$
are fixed across all held-out datasets.

\section{Experiments}
\label{expriments}

We test whether SWEET improves reasoning accuracy in the presence of irrelevant information while preserving accuracy on the original problems. We describe the experimental setup (Section~\ref{sec:experimental_settings}), then evaluate SWEET against baselines and ablated variants (Section~\ref{sec:main_results}). We also assess whether its gains hold across model families and scales (Section~\ref{sec:cross_model_robustness}).

\begin{table*}[t]
\centering
\resizebox{\textwidth}{!}{%
\begin{tabular}{lccccccccccccc}
\toprule
\multirow{2}{*}{Method}
& \multicolumn{2}{c}{GSM-Sym}
& \multicolumn{2}{c}{Math500}
& \multicolumn{2}{c}{AMC}
& \multicolumn{2}{c}{Gaokao-Math}
& \multicolumn{2}{c}{AIME}
& \multicolumn{3}{c}{Average} \\
\cmidrule(lr){2-3}\cmidrule(lr){4-5}\cmidrule(lr){6-7}
\cmidrule(lr){8-9}\cmidrule(lr){10-11}\cmidrule(lr){12-14}
& NoOp & Orig & NoOp & Orig & NoOp & Orig
& NoOp & Orig & NoOp & Orig & NoOp & Orig & Gap (\%) \\
\midrule

\multicolumn{14}{@{}l}{\textit{DeepSeek-R1-Distill-Qwen-32B}} \\
Base
& 80.1 & 87.5 & 79.8 & 85.8 & 67.7 & 77.2 & 62.3 & 71.9 & 42.2 & 54.4 & 66.4 & 75.4 & 11.9 \\
Steering
& 81.4 & 88.0 & 79.0 & 89.1 & 75.6 & 79.5 & 66.8 & 73.2 & 50.0 & 47.8 & 70.5 & 75.5 & 6.6 \\
ThinkEdit
& 79.4 & 88.0 & 79.0 & 87.2 & 66.1 & 74.8 & 65.2 & 72.7 & 42.2 & 45.6 & 66.4 & 73.6 & 9.9 \\
Steer2Edit
& 80.9 & 87.7 & 81.4 & 86.6 & 76.4 & 80.3 & 66.2 & 74.5 & 50.0 & 55.6 & 71.0 & 76.9 & 7.7 \\
SWEET (TL)
& 83.1 & 85.8 & \textbf{86.6} & 88.5 & 78.7 & 77.2 & 71.7 & 71.2 & 42.2 & 47.8 & 72.5 & 74.1 & \textbf{2.2} \\
SWEET (SL)
& 81.5 & 88.2 & 84.7 & 86.3 & 78.7 & 78.0 & 69.4 & 74.0 & 48.9 & 55.6 & 72.6 & 76.4 & 4.9 \\
SWEET (SL+TL)
& \textbf{84.8} & 87.1 & 85.8 & 88.0 & \textbf{82.7} & 85.0 & \textbf{74.3} & 74.8 & \textbf{51.1} & 55.6 & \textbf{75.7} & 78.1 & 3.0 \\

\midrule
\multicolumn{14}{@{}l}{\textit{DeepSeek-R1-Distill-Llama-8B}} \\
Base
& 58.1 & 73.5 & 59.5 & 78.9 & 47.2 & 63.0 & 46.8 & 61.0 & 18.9 & 23.3 & 46.1 & 60.0 & 23.1 \\
Steering
& 60.9 & 74.1 & 66.0 & 80.3 & 52.0 & 67.7 & 54.5 & 64.9 & 30.0 & 28.9 & 52.7 & 63.2 & 16.6 \\
ThinkEdit
& 61.8 & 74.9 & 69.9 & 81.9 & 50.4 & 66.9 & 55.8 & 65.5 & 30.0 & 30.0 & 53.6 & 63.8 & 16.1 \\
Steer2Edit
& 62.5 & 74.9 & 69.6 & 82.2 & 60.6 & 68.5 & 59.2 & 66.8 & 30.0 & 34.4 & 56.4 & 65.4 & 13.7 \\
SWEET (TL)
& \textbf{65.4} & 70.2 & 74.0 & 80.5 & 63.0 & 65.4 & 59.2 & 64.7 & 33.3 & 31.1 & 59.0 & 62.4 & 5.4 \\
SWEET (SL)
& 62.2 & 74.5 & 71.2 & 84.4 & 60.6 & 65.4 & 56.1 & 66.2 & 27.8 & 37.8 & 55.6 & 65.6 & 15.3 \\
SWEET (SL+TL)
& 64.9 & 71.3 & \textbf{77.5} & 82.2 & \textbf{66.9} & 66.9 & \textbf{63.6} & 64.9 & \textbf{34.4} & 31.1 & \textbf{61.5} & 63.3 & \textbf{2.9} \\
\bottomrule
\end{tabular}
}
\caption{Reasoning accuracy (\%) under original (Orig) and perturbed (NoOp) inputs. TL and SL denote TL-MLP25 and SL-Attn100, respectively; SL+TL combines both edits. Average reports the unweighted mean across the five benchmarks. Gap is $100\times|\mathrm{Orig}-\mathrm{NoOp}|/\mathrm{Orig}$, computed from unrounded average accuracies. Bold marks the highest NoOp accuracies and smallest Gap within each model.}
\label{tab:main-results}
\end{table*}

\begin{figure*}[t]
    \centering
    \includegraphics[width=0.98\textwidth]{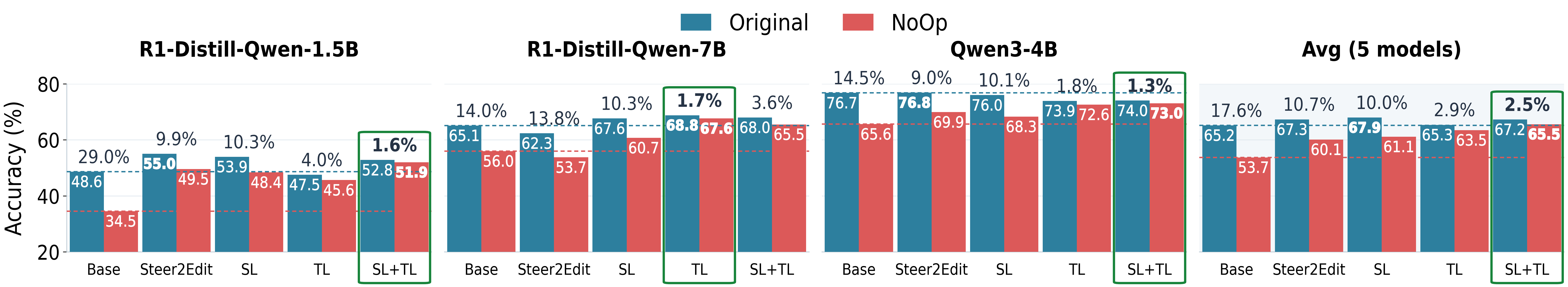}
        \caption{Original and perturbed (NoOp) accuracy across five reasoning benchmarks. The first three panels show DeepSeek-R1-Distill-Qwen-1.5B, DeepSeek-R1-Distill-Qwen-7B, and Qwen3-4B; the final panel averages all five models. Dashed lines mark the corresponding Base accuracies. TL and SL denote TL-MLP25 and SL-Attn100, respectively. Values above bar pairs report the relative accuracy gap, $100\times|\mathrm{Orig}-\mathrm{NoOp}|/\mathrm{Orig}$. Bold marks the highest Orig and NoOp accuracies and smallest gap within each panel; green boxes highlight the variant with the smallest gap. See Figure~\ref{fig:accuracy_gap_full} in Appendix~\ref{app:breakdown} for detailed results.}
    \label{fig:accuracy_gap}
\end{figure*}

\subsection{Settings}
\label{sec:experimental_settings}
\paragraph{Models and data.}
We evaluate five reasoning models: DeepSeek-R1-Distill-Qwen-1.5B, -7B and -32B,
DeepSeek-R1-Distill-Llama-8B~\citep{guo2025deepseek}, and Qwen3-4B~\citep{yang2025qwen3}.
We report per-dataset results for DeepSeek-R1-Distill-Qwen-32B and DeepSeek-R1-Distill-Llama-8B, and accuracy averaged
over the five benchmarks for the remaining models.
We use five reasoning benchmarks: Math500,
AMC~\citep{maa_amc},
Gaokao-Math~\citep{liao2024mario}, AIME~\citep{maa_aime}, and GSM-Symbolic~\citep{mirzadeh2025gsm}, each in its original and
perturbed (NoOp) form. Following~\citet{fu2026correlation},
NoOp variants are generated with GPT-5.4-mini.%

\paragraph{Baselines and variants.}
We compare SWEET with the unedited model (\emph{Base}), activation steering (\emph{Steering})~\citep{turner2023steering,rimsky2024steering}, ThinkEdit~\citep{sun2025thinkedit}, and Steer2Edit~\citep{sun2026steer2edit}.
We adapt ThinkEdit and Steer2Edit using sequence-level directions extracted from original--NoOp pairs, following their respective component-selection and weight-editing procedures.
For each method, we select tunable settings on Math500 original--NoOp pairs excluded from final evaluation.
Appendix~\ref{app:calibration} details the baseline adaptations, calibration subsets, candidate settings, and selection results.
Following the component assignment in Section~\ref{sec:sparsity_calibration}, SWEET (TL) edits MLP neurons, SWEET (SL) edits attention heads, and full SWEET (SL+TL) combines the two edits.

\paragraph{Inference and hyperparameters.}
We use vLLM~\citep{kwon2023efficient} with bfloat16 precision, greedy decoding, and a maximum of 16,384 generated tokens. Direction extraction uses 20 Math500 original--NoOp pairs per model. SWEET edits the top 25\% of positively attributed MLP neurons for TL and all positively attributed attention heads for SL. 
Calibration details and baseline hyperparameters are provided in
Appendix~\ref{app:calibration}.

\subsection{Main Results}
\label{sec:main_results}
Table~\ref{tab:main-results} shows that full SWEET~(SL+TL) achieves the highest average NoOp accuracy on both the 32B and 8B models ($75.7\%$ and $61.5\%$), outperforming the strongest baseline, Steer2Edit, by $4.7$ and $5.1$ percentage points, respectively. On the 32B model, all three SWEET variants outperform every baseline in average NoOp accuracy ($72.5$--$75.7\%$ vs.\ at most $71.0\%$) and achieve smaller relative Orig--NoOp gaps ($2.2$--$4.9\%$ vs.\ at least $6.6\%$). TL and SL+TL also outperform all baselines on both metrics on the 8B model. Full SWEET additionally achieves the highest average Orig accuracy on the 32B model ($78.1\%$), while SL achieves the highest on the 8B model ($65.6\%$). These results demonstrate that SWEET improves reasoning robustness to irrelevant distractors while maintaining or improving average accuracy on original problems.

\paragraph{Joint editing improves NoOp accuracy over either direction alone.}
Full SWEET improves average NoOp accuracy over the stronger single-direction variant by $3.1$ and $2.5$ percentage points on the 32B and 8B models, respectively. The relative effectiveness of TL and SL varies across models: they achieve similar NoOp accuracy on the 32B model ($72.5\%$ and $72.6\%$, respectively), whereas TL outperforms SL on the 8B model ($59.0\%$ vs.\ $55.6\%$). SL achieves higher average Orig accuracy than TL on both models. Joint editing surpasses SL in Orig accuracy on the 32B model ($78.1\%$ vs.\ $76.4\%$), but falls below it on the 8B model ($63.3\%$ vs.\ $65.6\%$). Thus, combining the directions consistently improves NoOp accuracy, although its benefits for Orig accuracy vary across models.

\paragraph{SWEET reduces the original--perturbed accuracy gap.}
Compared with Base, full SWEET reduces the relative gap between domain-averaged Orig and NoOp accuracies from $11.9\%$ to $3.0\%$ on the 32B model and from $23.1\%$ to $2.9\%$ on the 8B model. Steer2Edit achieves gaps of $7.7\%$ and $13.7\%$, respectively. On the 32B model, TL achieves a smaller gap ($2.2\%$) but lower Orig and NoOp accuracies than full SWEET, illustrating why gap reduction should be assessed alongside absolute accuracy. Full SWEET improves both average accuracies over Base on both models, although trade-offs remain on individual benchmarks: Llama-8B's Orig accuracy on GSM-Symbolic declines from $73.5\%$ to $71.3\%$.

\subsection{Robustness across Model Families and Scales}
\label{sec:cross_model_robustness}
Figure~\ref{fig:accuracy_gap} reports results for the remaining three reasoning models and averages across all five. Full per-dataset results for all models are provided in Table~\ref{tab:main-results-full} in Appendix~\ref{app:breakdown}. Averaged across all five models, all three SWEET variants outperform the compared baselines in average NoOp accuracy ($61.1$--$65.5\%$ vs.\ at most $60.1\%$) and achieve smaller relative Orig--NoOp gaps ($2.5$--$10.0\%$ vs.\ at least $10.7\%$). Full SWEET achieves the highest average NoOp accuracy ($65.5\%$) and smallest relative gap ($2.5\%$). It also improves average Orig accuracy over Base from $65.2\%$ to $67.2\%$, close to SL ($67.9\%$) and Steer2Edit ($67.3\%$). The results demonstrate SWEET's effectiveness across Qwen2.5, Qwen3, and Llama3.1 models ranging from 1.5B to 32B parameters.

\section{Analysis}
\label{analysis}

\begin{table}[t]
\centering
\scriptsize
\setlength{\tabcolsep}{1pt}
\renewcommand{\arraystretch}{1.2}
\begin{tabular}{@{}lp{0.80\columnwidth}@{}}
\toprule
\shortstack[l]{Direction\\Model} & Top-ranked tokens \\
\midrule
TL R1-1.5B &
\texttt{\textvisiblespace{}}\textbf{\texttt{also}},\hspace{0.08em}\texttt{\textvisiblespace{}}\textbf{\texttt{given}},\hspace{0.08em}\texttt{\textvisiblespace{}}\textbf{\texttt{mention}},\hspace{0.08em}\texttt{\textvisiblespace{}}\textbf{\texttt{additionally}}\\
TL R1-7B &
\texttt{\textvisiblespace{}}\textbf{\texttt{additionally}},\hspace{0.08em}\textbf{\texttt{Additionally}},\hspace{0.08em}\texttt{\textvisiblespace{}}\textbf{\texttt{unrelated}},\hspace{0.08em}\texttt{\textvisiblespace{}}\textbf{\texttt{Additionally}}\\
TL R1-8B &
\textbf{\texttt{given}},\hspace{0.08em}\texttt{\textvisiblespace{}}\textbf{\texttt{given}},\hspace{0.08em}\texttt{\textvisiblespace{}}\textbf{\texttt{Given}},\hspace{0.08em}\textbf{\texttt{Given}}\\
TL R1-32B &
\textbf{\texttt{\begin{CJK}{UTF8}{gbsn}另有\end{CJK}}} \textbf{(there is also)},\hspace{0.08em}\textbf{\texttt{\begin{CJK}{UTF8}{gbsn}无关\end{CJK}}} \textbf{(unrelated)},\hspace{0.08em}\textbf{\texttt{\begin{CJK}{UTF8}{gbsn}另\end{CJK}}} \textbf{(other)},\hspace{0.08em}\textbf{\texttt{\begin{CJK}{UTF8}{gbsn}有个\end{CJK}}} \textbf{(there is a)}\\
TL Qwen3-4B &
\textbf{\texttt{\begin{CJK}{UTF8}{gbsn}还有一个\end{CJK}}} \textbf{(one more)},\hspace{0.08em}\texttt{\textvisiblespace{}}\textbf{\texttt{given}},\hspace{0.08em}\textbf{\texttt{\begin{CJK}{UTF8}{gbsn}提及\end{CJK}}} \textbf{(mention)},\hspace{0.08em}\textbf{\texttt{\begin{CJK}{UTF8}{gbsn}提到\end{CJK}}} \textbf{(mention)}\\
\addlinespace[2pt]
SL R1-1.5B &
\texttt{OCR},\hspace{0.08em}\texttt{ird},\hspace{0.08em}\texttt{\textvisiblespace{}\$\#},\hspace{0.08em}\texttt{\textvisiblespace{}}\textbf{\texttt{nevertheless}}\\
SL R1-7B &
\texttt{ieber},\hspace{0.08em}\texttt{allet},\hspace{0.08em}\texttt{\textvisiblespace{}asn},\hspace{0.08em}\texttt{ITU}\\
SL R1-8B &
\texttt{\textvisiblespace{}Cor},\hspace{0.08em}\texttt{idebar},\hspace{0.08em}\texttt{anche},\hspace{0.08em}\texttt{\textvisiblespace{}Devin}\\
SL R1-32B &
\texttt{arend},\hspace{0.08em}\texttt{/swagger},\hspace{0.08em}\texttt{\textvisiblespace{}Berg},\hspace{0.08em}\textbf{\texttt{\begin{CJK}{UTF8}{gbsn}无论如何\end{CJK}}} \textbf{(in any case)}\\
SL Qwen3-4B &
\texttt{icorn},\hspace{0.08em}\texttt{\begin{CJK}{UTF8}{gbsn}特别声明\end{CJK}} (special statement),\hspace{0.08em}\texttt{eyJ},\hspace{0.08em}\textbf{\texttt{\begin{CJK}{UTF8}{gbsn}某种\end{CJK}}} \textbf{(some kind of)}\\
\bottomrule
\end{tabular}
\caption{Top four tokens, in rank order, from vocabulary readouts at 75\% of model depth. \texttt{\textvisiblespace} denotes a literal space; capitalization is preserved, and tokens with distinct IDs are retained separately. Bold highlights selected interpretable tokens, and parenthetical English glosses aid interpretation; neither affects the ranking. Appendix~\ref{app:vocabulary-readout} provides the procedure, complete top-10 lists, token IDs, and additional readouts at 25\% and 50\% depth to assess sensitivity to layer choice.}
\label{tab:tltokens}
\end{table}

\paragraph{Interpreting behavioral directions through vocabulary projections.}
Following Appendix~\ref{app:vocabulary-readout}, we project the extracted directions through the unembedding matrix and rank tokens by their mean raw projection scores (Table~\ref{tab:tltokens}). TL readouts include additive and premise-introducing terms such as \emph{also}, \emph{additionally}, \emph{given}, \emph{mention}, and their Chinese counterparts. Since the perturbations introduce irrelevant conditions, these readouts motivate the hypothesis that TL captures their incorporation into reasoning, which we term \emph{over-integration}. SL readouts are noisier and less interpretable. Logit-lens interpretations can be brittle~\citep{belrose2023tunedlens}, particularly for SL directions, which contrast mean-pooled states rather than states at a shared next-token position. We therefore treat the link between SL and overthinking as a behavioral hypothesis and test it below.

\paragraph{TL and SL have complementary effects on reasoning robustness.}

We compare SL, TL, and SL+TL on original and NoOp inputs, assessing over-integration and overthinking separately (Table~\ref{tab:analysis-summary}). For over-integration, a blinded GPT-6-Luna judge compares Base and edited traces for the same NoOp problem in both presentation orders. We count a win (loss) when both orders indicate less (more) over-integration in the edited trace, reporting rates over all evaluated trace pairs. We complement these judgements with lexical reuse: the relative change in the fraction of responses that repeat wording from the irrelevant condition, detected through rule-based 3-gram matching. Appendix~\ref{app:behavioral-judgments} provides further details and discussion of these measures. We assess overthinking through generation length, token efficiency, token-cap incidence~\citep{sfwe_ot_kaiser2026}, and repetition-loop incidence in length-matched prefixes~\citep{sfwe_ot_alimaskina2026}, with metric definitions in Appendix~\ref{app:saved-trace-efficiency}.

TL reduces both over-integration and overthinking on NoOp inputs. Its over-integration win/loss rates are $51.6\%/4.4\%$, compared with $26.4\%/20.2\%$ for SL, indicating a stronger reduction in over-integration. Lexical reuse shows a corresponding difference: TL reduces the fraction of responses repeating distractor wording by $77.5\%$, compared with $0.6\%$ for SL. Other LLM judges support the same qualitative conclusion (Appendix~\ref{app:cross-judge-robustness}). On original inputs, where no irrelevant condition is added, TL's effect on overthinking is much weaker: token efficiency changes by $-0.5\%$, compared with $+56.1\%$ on NoOp, and reductions in generation length and token-cap incidence are smaller than for SL. This pattern suggests that TL primarily targets over-integration and that its reduction of overthinking on NoOp may partly reflect reduced incorporation of distractors. Attention analysis further shows that TL-active positions attend strongly to distractors (Appendix~\ref{app:attention-analysis}).

SL reduces overthinking on both original and NoOp inputs, with token-efficiency gains of $10.4\%$ and $26.6\%$, respectively. Although its NoOp gains are smaller than TL's, its benefits on original inputs support the hypothesis that SL addresses broader overthinking tendencies. Combining the directions gives the strongest improvements across all measured metrics on NoOp and the largest reductions in generation length, token-cap incidence, and repetition loops on original inputs. These results support combining TL and SL to suppress both forms of spurious reasoning.

\paragraph{SWEET's robustness gains extend beyond mitigating overthinking.}
To assess whether reducing overthinking alone can match SWEET's robustness gains, we compare SWEET with budget forcing~\citep{muennighoff2025s1}, DEER~\citep{yang2026dynamic}, SEAL~\citep{chen2025seal}, and Chain-of-Draft~\citep{xu2025chain}. SWEET achieves the highest NoOp accuracy ($65.54\%$; Table~\ref{tab:overthinking-baselines-full}), exceeding SEAL, the strongest overthinking baseline, by $2.84$ percentage points while maintaining comparable Orig accuracy ($67.23\%$ vs.\ $67.84\%$). Its relative gap is also smaller than SEAL's ($2.52\%$ vs.\ $7.57\%$). Chain-of-Draft achieves a slightly smaller gap ($2.36\%$), but lowers both Orig and NoOp accuracy below Base. All four baselines reduce NoOp token counts more than SWEET ($27.0$--$46.5\%$ vs.\ $23.3\%$), yet none matches its NoOp accuracy. These results suggest that shortening outputs alone is insufficient to match SWEET's gains. %

\paragraph{Additional ablations and sensitivity analyses.}
We assess direction stability across different samples and extraction sizes (Appendix~\ref{app:native-sl-complete}). We also examine sensitivity to editing sparsity and hyperparameter choices (Appendix~\ref{app:calibration}) and compare alternative component assignments (Appendix~\ref{app:breakdown}).

\begin{table}[t]
\centering
\small
\setlength{\tabcolsep}{1.0pt}
\renewcommand{\arraystretch}{1.1}
\begin{tabular}{@{}llcccccc@{}}
\toprule
& & \multicolumn{2}{c}{Over-integr.}
& \multicolumn{4}{c}{Overthinking} \\
\cmidrule(lr){3-4}
\cmidrule(lr){5-8}
Input & Edit
& \shortstack{Judge W/L\\(\%) $\uparrow/\downarrow$}
& \shortstack{Reuse\\(\%) $\downarrow$}
& \shortstack{Tokens\\(\%) $\downarrow$}
& \shortstack{Tok. Eff.\\(\%) $\uparrow$}
& \shortstack{Cap\\(\%) $\downarrow$}
& \shortstack{Loop\\(pp) $\downarrow$} \\
\midrule
Orig & SL
& -- & -- & $-4.3$ & $+10.4$ & $-21.3$ & $-0.90$ \\
& TL
& -- & -- & $-3.5$ & $-0.5$ & $-12.3$ & $-2.11$ \\
& Both
& -- & -- & $-6.4$ & $+6.2$ & $-28.7$ & $-3.53$ \\
\midrule
NoOp & SL
& 26.4 / 20.2 & $-0.6$
& $-10.2$ & $+26.6$ & $-31.9$ & $-3.00$ \\
& TL
& 51.6 / 4.4 & $-77.5$
& $-20.0$ & $+56.1$ & $-33.8$ & $-6.29$ \\
& Both
& 52.8 / 3.4 & $-78.8$
& $-23.3$ & $+65.8$ & $-47.6$ & $-7.28$ \\
\bottomrule
\end{tabular}
\caption{Changes in over-integration and overthinking under SL, TL,
and combined editing relative to Base.
Judge W/L reports the percentages of all evaluated NoOp trace pairs
showing less/more over-integration after editing, requiring agreement
across both presentation orders.
Reuse reports the relative change (\%) in the fraction of responses
containing lexical reuse of the irrelevant condition, detected by
3-gram matching.
Tokens, token efficiency, and Cap report relative changes (\%)
in output length, token efficiency, and token-cap incidence,
respectively.
Loop reports the percentage-point change in repetition-loop incidence
over length-matched prefixes.
``--'' denotes not applicable.}
\label{tab:analysis-summary}
\end{table}

\begin{table}[t]
\centering
\small
\setlength{\tabcolsep}{2pt}
\begin{tabular*}{\columnwidth}{@{\extracolsep{\fill}}lrrrr@{}}
\toprule
Method & Orig. (\%) & NoOp (\%) & Gap (\%) & $\Delta$Tok. (\%) \\
\midrule
Base           & 65.16 & 53.72 & 17.55 & ---   \\
Budget forcing & 65.29 & 62.42 &  4.39 & $-46.5$ \\
DEER           & 66.38 & 62.58 &  5.72 & $-36.7$ \\
SEAL           & \textbf{67.84} & 62.70 & 7.57 & $-27.0$ \\
Chain-of-Draft & 52.18 & 50.95 &  2.36 & $-39.5$ \\
\midrule
SWEET          & 67.23 & \textbf{65.54} & 2.52 & $-23.3$ \\
\bottomrule
\end{tabular*}
\caption{Comparison of SWEET with methods for reducing overthinking. Orig and NoOp report accuracy (\%) on original and distractor-augmented inputs, averaged equally across five benchmarks and five models; Gap is their relative accuracy gap. $\Delta$Tok.\ reports the percentage change in macro-averaged NoOp API completion-token counts relative to Base, with negative values indicating shorter outputs (Appendix~\ref{app:saved-trace-efficiency}). Bold marks the highest accuracy in each input condition.}
\label{tab:overthinking-baselines-full}
\end{table}

\section{Related Work}
\label{related_work}
\paragraph{Mitigating spurious correlations.}
Work on spurious correlations has predominantly focused on natural language understanding, using ensembles~\citep{mahabadi2020end}, token masking~\citep{wang2022identifying}, data augmentation~\citep{DBLP:conf/acl/Korakakis0W25}, causal reweighting~\citep{zhou2025fighting}, or neuron pruning based on Integrated Gradients~\citep{ali2025detecting}. These methods target shortcuts in classification, leaving unclear how to identify and suppress spurious dependencies throughout long CoT reasoning. Reasoning-specific methods address this challenge through training or reasoning-path selection: LeaF combines confounding-token pruning with attention distillation~\citep{guo2026learning}, DeCoT uses causal interventions to select reasoning paths~\citep{wu2024decot}, and \citet{xie2025mitigating} use correctness-perception distillation to reduce spurious question--answer correlations. SWEET intervenes directly in the model's internal computation: it identifies attention heads and MLP neurons associated with spurious-feature dependence and edits their weights to suppress their contributions---without retraining the model or adding reasoning-path selection at inference time.

\paragraph{Analyzing and modifying model behavior.}
Activation steering~\citep{turner2023steering,rimsky2024steering} and parameter interventions have been used to study and alter model behavior, including safety~\citep{wei2024assessing} and CoT faithfulness~\citep{tutek2025measuring}. ThinkEdit~\citep{sun2025thinkedit} and Steer2Edit~\citep{sun2026steer2edit} perform component-level weight editing using directions derived from sequence-averaged activation contrasts. In contrast to these methods, SWEET combines answer-preserving counterfactual interventions with token- and sequence-level extraction to distinguish local responses to irrelevant information from changes across the reasoning trajectory. Token-level extraction compares activations along the same reasoning trace under the original and perturbed prompts, whereas sequence-level extraction compares mean activations across independently generated traces. Together, these contrasts identify components associated with distinct forms of spurious reasoning, and guide weight edits that suppress their contributions.

\section{Conclusion}
\label{conclusion}
We introduced SWEET, a training-free method that combines mechanistic interpretability with counterfactual interventions to identify spurious-feature reliance in CoT reasoning at both token and sequence levels and suppress it through weight edits. Across five open-weight reasoning models and five benchmarks, SWEET improves average accuracy on both original and perturbed inputs while substantially reducing the relative accuracy gap between them. Detailed analyses provide mechanistic evidence that these robustness gains stem from suppressing two recurring forms of spurious reasoning: \emph{over-integration} and \emph{overthinking}.

\section*{Limitations}
Our study has three limitations. First, although we use widely adopted benchmarks, these provide only a partial assessment of reasoning robustness: our evaluation focuses on irrelevant-information perturbations in mathematical reasoning, leaving applicability to other tasks and perturbation types unverified. Second, mitigating spurious-feature reliance can involve trade-offs between robustness and task accuracy. Although SWEET improves accuracy on both original and perturbed inputs on average, editing can reduce performance on some model--dataset combinations. Further work is needed to understand when these trade-offs arise and which capabilities are affected. Third, SWEET requires calibration data and additional computation to select editing sparsities, although this procedure is less computationally demanding than training-based mitigation. We use a common selection procedure across models, keep the selected settings fixed across held-out datasets, and demonstrate robustness across the tested sparsity settings. Sensitivity to other calibration distributions and unexplored parameter ranges remains unverified.

\bibliography{custom}

\appendix
\clearpage
\section*{Appendix Overview}
The appendix first specifies calibration and evaluation settings, then gives
complete accuracy results and baseline comparisons. It then presents attention
analysis, the behavioral-judging protocol, and definitions of the automatic
metrics used in Table~\ref{tab:analysis-summary}, followed by the vocabulary-readout procedure and depth-sensitivity lists.
Throughout, \emph{clean} and \emph{original} refer to the same unperturbed input
condition.

\section{Hyperparameter Calibration and Evaluation Settings}
\label{app:calibration}
\label{sec:calibration_results}
This section specifies calibration data, candidate settings, and evaluation
protocols for SWEET, Steering, ThinkEdit, and Steer2Edit. It also reports
direction stability across extraction sizes and component sparsity sweeps.
Calibration examples are excluded from final evaluation. 
All methods draw their calibration data from the same pool of 100
Math500 problems and their NoOp counterparts. Within each calibration
comparison, all candidate settings are evaluated on identical question
IDs under both input conditions, with fitting-example exclusions
specified below, ensuring matched comparisons among candidates.

\paragraph{Direction-extraction settings.}
For both TL and SL, we set $\lambda=0.25$ in the variance-normalized
estimator (Equation~\ref{eq:variance_normalisation}).
For TL, response positions are ranked by their tokenwise KL divergence;
we retain the smallest set whose cumulative divergence reaches at least
50\% of the total for each reference trace. Thus, the threshold concerns
KL mass rather than the fraction of tokens retained.
These settings are fixed across models.

\paragraph{SWEET calibration and evaluated settings.}
Following Section~\ref{sec:sparsity_calibration}, the main comparisons
assign TL edits to MLP neurons and SL edits to attention heads.
We calibrate the editing fraction within each component type on the first
100 Math500 problems and their NoOp counterparts, using mean Orig and NoOp
accuracy averaged across the five models.
The candidate fractions are
$\tau\in\{0.25,1,2,5,10,25,100\}\%$ for attention heads and
$\tau\in\{0.01,0.05,0.25,1,5,25,100\}\%$ for MLP neurons.
Fractions refer to positively attributed units within each direction and
component type, pooled across layers.

The main comparisons use TL-MLP25 and SL-Attn100; full SWEET combines
these edits. Table~\ref{tab:regret} summarizes the calibration results.
TL-MLP achieves its minimum mean regret at 25\%. SL-head budgets of
25\% and 100\% tie at a five-model calibration accuracy of $81.5\%$;
we use 100\% in all main comparisons.
The component-variant comparison below evaluates MLP, Attn, and Both
at fixed budgets. Its test results are reported in
Appendix~\ref{app:breakdown}.

\paragraph{Component-variant calibration.}
We compare MLP, Attn, and Both variants separately for TL and SL.
The positive-unit budgets are 25\% for TL neurons and heads,
5\% for SL neurons, and 100\% for SL heads. Both applies the
two budgets of its direction; no additional joint sparsity search
is performed. Table~\ref{tab:component-variant-calibration} reports
mean Orig and NoOp accuracy, with equal weight across five models,
on Math500 IDs 0--99. Since these 100 pairs include direction-fitting
examples, we also report the common 87-pair subset excluding SWEET
extraction examples and the common 85-pair subset used for ThinkEdit
and Steer2Edit calibration.

TL-MLP25 and SL-MLP5 achieve the highest calibration accuracy within
their respective directions on all three subsets. This comparison
assesses alternative single-direction component assignments. The
main-table TL and SL ablations instead use the components of the fixed
SWEET composition, TL-MLP25 and SL-Attn100, respectively.

\begin{table*}[t]
\centering
\small
\caption{Component-variant calibration accuracy (\%) averaged equally
over Orig/NoOp and five models. Budgets are percentages of positively
attributed units; -- means that component type is unedited.
First100 uses Math500 IDs 0--99, Excl87 excludes SWEET extraction
examples, and Common85 uses the common fitting-excluded baseline
calibration subset. Bold marks the highest accuracy within each
direction and subset.}
\label{tab:component-variant-calibration}
\begin{tabular}{lrrrrr}
\toprule
Variant & Neurons (\%) & Heads (\%) & First100 & Excl87 & Common85 \\
\midrule
TL-MLP  & 25 & --  & \textbf{83.30} & \textbf{84.60} & \textbf{84.47} \\
TL-Attn & -- & 25  & 81.50 & 82.41 & 82.35 \\
TL-Both & 25 & 25  & 82.50 & 83.10 & 82.94 \\
\midrule
SL-MLP  & 5  & --  & \textbf{82.50} & \textbf{84.14} & \textbf{84.00} \\
SL-Attn & -- & 100 & 81.50 & 82.64 & 82.71 \\
SL-Both & 5  & 100 & 81.00 & 82.76 & 82.82 \\
\bottomrule
\end{tabular}
\end{table*}

\paragraph{Steering hyperparameter selection.}
We test TL, SL, and their combination with steering
coefficients $c\in\{0.1,0.5,1\}$ on the first 100 Math500
problems and their NoOp variants. For combined steering,
we sum the separately scaled TL and SL vectors and apply
the same coefficient to both. Table~\ref{tab:steering-calibration}
reports original and NoOp accuracy, each averaged equally
across the five models. SL with $c=0.1$ achieves the highest
mean accuracy across the two input conditions (82.7\%)
and is used in the reported evaluations.

\begin{table}[t]
\centering
\small
\begin{tabular}{lrrrr}
\toprule
Direction & $c$ & Orig & NoOp & Mean \\
\midrule
TL & 0.1 & 85.0 & 67.6 & 76.3 \\
TL & 0.5 & 84.8 & 70.4 & 77.6 \\
TL & 1.0 & 84.6 & 69.8 & 77.2 \\
\midrule
SL & 0.1 & 88.0 & \textbf{77.4} & \textbf{82.7} \\
SL & 0.5 & 87.0 & \textbf{77.4} & 82.2 \\
SL & 1.0 & 86.8 & 75.8 & 81.3 \\
\midrule
TL+SL & 0.1 & \textbf{88.2} & 76.2 & 82.2 \\
TL+SL & 0.5 & 87.0 & 76.8 & 81.9 \\
TL+SL & 1.0 & 87.8 & 75.8 & 81.8 \\
\bottomrule
\end{tabular}
\caption{Steering calibration accuracy (\%) on the first 100
Math500 problems in original and NoOp form. Orig and NoOp weight the
five models equally; Mean averages the two conditions and selects SL
with $c=0.1$. Bold marks the highest score in each column.}
\label{tab:steering-calibration}
\end{table}

\paragraph{ThinkEdit hyperparameter selection.}
Following ThinkEdit's sequence-level formulation, we use
sequence-level directions extracted from our original--perturbed
pairs. With these directions and global head rankings fixed,
we test editing 0, 0.25, 1, 2, 4, 5, 8, 10, 25, and 100\%
of all attention heads across layers, rounding the head count
to the nearest integer. We evaluate each setting on the first
100 Math500 problems and their NoOp variants, and use the common
85 pairs remaining after excluding known direction-fitting examples
to choose one shared percentage across models.
Table~\ref{tab:thinkedit-calibration} reports Orig and NoOp accuracy,
each averaged equally across the five models.
Editing 5\% of heads gives the highest mean accuracy across
the two input conditions. This corresponds to 17, 39, 51, 128,
and 58 heads for DS1.5B, DS7B, Llama8B, DS32B, and Qwen3-4B,
respectively, and is used for all test datasets.

\begin{table}[t]
\centering
\small
\begin{tabular}{lrrr}
\toprule
Head budget & Orig & NoOp & Mean \\
\midrule
0\%    & 86.59 & 68.00 & 77.29 \\
0.25\% & 87.06 & 71.29 & 79.18 \\
1\%    & 85.65 & 71.76 & 78.71 \\
2\%    & 86.12 & 70.59 & 78.35 \\
4\%    & 85.65 & 75.06 & 80.35 \\
5\%    & 86.35 & \textbf{76.71} & \textbf{81.53} \\
8\%    & \textbf{87.53} & 72.94 & 80.24 \\
10\%   & 86.82 & 72.94 & 79.88 \\
25\%   & 87.06 & 66.59 & 76.82 \\
100\%  & 69.18 & 60.24 & 64.71 \\
\bottomrule
\end{tabular}
\caption{ThinkEdit calibration accuracy (\%) on the common 85
Math500 pairs after excluding known fitting examples.
Orig and NoOp weight the five models equally; Mean averages the
two conditions and selects the shared 5\% head budget.
Percentages refer to all attention heads, retaining ThinkEdit's
signed global ranking; SWEET's sparsity percentages instead refer
to positively attributed units. Bold marks the highest score
in each accuracy column.}
\label{tab:thinkedit-calibration}
\end{table}

\paragraph{Evaluation details.}
Final evaluation excludes extraction and calibration data, as well as the
compared baselines' direction- and scale-fitting data.
Generation stops at the 16,384-token limit even when a
response is incomplete.

\begin{table*}[t]
\centering
\small
\setlength{\tabcolsep}{6pt}
\renewcommand{\arraystretch}{1.05}
\begin{SWEETfitwidth}
\begin{tabular}{rrrrr}
\toprule
& \multicolumn{2}{c}{MLP neurons} & \multicolumn{2}{c}{Attention heads} \\
\cmidrule(lr){2-3} \cmidrule(lr){4-5}
$\tau$ (\%) & TL--MLP & SL--MLP & TL--Attn & SL--Attn \\
\midrule
0.01  & 6.20 & 6.70 & --   & --   \\
0.05  & 6.80 & 6.40 & --   & --   \\
0.25  & 5.90 & 3.70 & 5.80 & 8.10 \\
1     & 4.30 & 3.30 & 5.10 & 6.70 \\
2     & --   & --   & 4.50 & 6.70 \\
5     & 1.70 & \textbf{2.00} & 3.80 & 4.00 \\
10    & --   & --   & 2.70 & 2.60 \\
25    & \textbf{0.80} & 3.80 & \textbf{0.80} & \textbf{1.30} \\
100   & 2.30 & 4.20 & 1.00 & \textbf{1.30} \\
\bottomrule
\end{tabular}
\end{SWEETfitwidth}
\caption{Mean regret $R_{g,m}(\tau)$ in accuracy points
(Equation~\ref{eq:regret}) on the first 100 Math500 problems and their
NoOp variants, with equal weight across the five models. Lower is better;
bold marks the minimum within each direction and unit type. Percentages
refer to positively attributed units.}
\label{tab:regret}
\end{table*}

\subsection{Steer2Edit Adaptation and Calibration}
\label{app:steer2edit}
\paragraph{Task adaptation and direction.}
We adapt Steer2Edit~\citep{sun2026steer2edit} to irrelevant-information
robustness, retaining its signed component scores, elastic-net coefficients,
and additive rank-one weight updates.
We use the SL direction estimated from the same 20
original-correct/NoOp-incorrect pairs as SWEET, with the variance-normalized
estimator in Section~\ref{sec:direction_extraction}.
Negating the NoOp-ward unit direction gives the desired original-ward
direction $v_\ell$, shared by attention and MLP components at layer $\ell$.
Thus, we evaluate Steer2Edit's editing rule with a shared task-specific
direction rather than its original behavior-specific extraction pipeline.

\paragraph{Component statistics and editing.}
We replay the saved Orig and NoOp extraction traces on the unedited model
and collect inputs to attention output projections and MLP down projections.
Inputs are averaged over positions predicting response tokens within each
trace, then equally across examples and conditions to obtain $\bar{x}_i$.
Replay is right-truncated at 18,000 tokens, and statistics are shared across
all candidate configurations.
For component $i$, $W_i$ is an attention-head output-projection block or
an MLP-neuron output column. Its signed score and editing coefficient are
\begin{align}
g_i &= \cos(v_\ell,W_i\bar{x}_i),\\
\beta_i &=
\frac{\operatorname{sign}(g_i)\max(|g_i|-\rho_i\alpha,0)}
{\rho_i(1-\alpha)}.
\end{align}
Here $\rho_i$ controls regularization strength and $\alpha$ sets the
elastic-net mixture. With
$k_i=W_i^\top v_\ell/\lVert W_i^\top v_\ell\rVert_2$, the update is
\begin{equation}
W'_i=W_i+\beta_i v_\ell k_i^\top.
\end{equation}
For an MLP neuron, $k_i$ reduces to the sign of the scalar
$W_i^\top v_\ell$. Components with zero coefficient or zero
$W_i^\top v_\ell$ are unchanged.
Both positive and negative scores enter selection.
Each candidate starts from unedited weights; factors are computed in
float32 from bfloat16 weights, and updates are applied in bfloat16.

\paragraph{Calibration.}
We evaluate 18 configurations:
\[
\{\mathrm{Attn},\mathrm{MLP},\mathrm{Both}\}
\times\{0.3,0.5,0.7\}
\times\{0.1,0.3\},
\]
where the last two sets specify $\rho$ and $\alpha$.
Both uses the same $\rho$ for attention and MLP components;
independent regularization strengths for the two families are not searched.

All configurations are evaluated on Math500 IDs 0--99 in Orig and NoOp form.
Selection uses the same common 85 pairs as ThinkEdit, after excluding known
direction-fitting examples, and maximizes mean accuracy across conditions
and models. The excluded zero-based IDs are
14, 18, 21, 24, 29, 32, 34, 39, 44, 50, 58, 66, 87, 93, and 98.
Scores on all 100 pairs are reported as a sensitivity check.
Table~\ref{tab:steer2edit-calibration} selects Both with
$\rho_{\mathrm{attn}}=\rho_{\mathrm{mlp}}=0.5$ and $\alpha=0.1$,
the unique maximum at $83.65\%$ on the 85-pair subset.
The same setting ranks first on all 100 pairs ($81.90\%$).
It is fixed across models and test benchmarks; each model has its own
direction and component statistics.

\begin{table*}[t]
\centering
\small
\caption{Steer2Edit calibration accuracy (\%) for all 18 configurations.
Orig and NoOp average equally across five models on the common 85 pairs;
their mean determines selection. First100 reports mean accuracy on all
100 pairs as a sensitivity check. Bold marks the highest mean for each
subset. Both uses the listed $\rho$ for both component families.}
\label{tab:steer2edit-calibration}
\begin{tabular}{lrrrrrr}
\toprule
Components & $\rho$ & $\alpha$ & Orig & NoOp & Mean & First100 \\
\midrule
Attn & 0.3 & 0.1 & 87.53 & 76.47 & 82.00 & 80.20 \\
MLP & 0.3 & 0.1 & 67.29 & 50.35 & 58.82 & 57.80 \\
Both & 0.3 & 0.1 & 68.94 & 52.71 & 60.82 & 59.70 \\
Attn & 0.5 & 0.1 & 84.94 & 74.35 & 79.65 & 78.20 \\
MLP & 0.5 & 0.1 & 84.94 & 73.88 & 79.41 & 77.80 \\
Both & 0.5 & 0.1 & 89.18 & 78.12 & \textbf{83.65} & \textbf{81.90} \\
Attn & 0.7 & 0.1 & 86.35 & 70.59 & 78.47 & 77.50 \\
MLP & 0.7 & 0.1 & 85.88 & 73.18 & 79.53 & 78.90 \\
Both & 0.7 & 0.1 & 88.94 & 73.88 & 81.41 & 79.60 \\
Attn & 0.3 & 0.3 & 85.41 & 73.41 & 79.41 & 78.00 \\
MLP & 0.3 & 0.3 & 86.35 & 72.47 & 79.41 & 77.70 \\
Both & 0.3 & 0.3 & 85.88 & 71.29 & 78.59 & 77.30 \\
Attn & 0.5 & 0.3 & 85.65 & 70.82 & 78.24 & 76.70 \\
MLP & 0.5 & 0.3 & 84.94 & 67.06 & 76.00 & 75.50 \\
Both & 0.5 & 0.3 & 85.88 & 70.82 & 78.35 & 76.80 \\
Attn & 0.7 & 0.3 & 83.29 & 71.06 & 77.18 & 76.40 \\
MLP & 0.7 & 0.3 & 86.59 & 70.59 & 78.59 & 77.50 \\
Both & 0.7 & 0.3 & 84.47 & 68.00 & 76.24 & 75.50 \\
\bottomrule
\end{tabular}
\end{table*}

\paragraph{Test evaluation.}
The selected configuration uses greedy decoding and a 16,384-token
generation limit on the same common test IDs as the main comparisons.
Evaluation covers 19,674 responses across five models, five benchmarks,
and two input conditions. Configuration selection uses calibration
accuracy only.

\subsection{SL and TL extraction-size sensitivity}
\label{app:native-sl-complete}
\paragraph{SL extraction-size sensitivity.}
We assess how the number of original--NoOp response pairs used for SL
extraction affects agreement with a direction estimated from a disjoint
reference set. All five models use saved own-response pairs that were answered
correctly under the original prompt and incorrectly under NoOp. The eligible
pools contain 144, 102, 130, 66, and 76 pairs for DS1.5B, DS7B, Llama8B,
DS32B, and Qwen3-4B, respectively; the DS7B pool is the verified
102-pair baseline-eligible intersection.

For each of 64 seeded splits, all extraction sizes share a fixed disjoint
reference within the model. Reference sets contain 26 pairs, except for
DS32B, which uses 16. Directions use the paired-effect estimator with
$\lambda=0.25$, an 18,000-input-token replay cap, and native chat tokenization
without an extra BOS token. Table~\ref{tab:native-sl-five} reports mean
layerwise direction cosine and the 2.5th--97.5th percentiles across the
finite-pool splits. These are descriptive split percentiles, not population
confidence intervals. Because reference sizes differ across models, the
comparison primarily measures within-model sensitivity.

\begin{table*}[t]
\centering\small
\begin{SWEETfitwidth}
\begin{tabular}{lccc}
\toprule Model & $n=20$ & $n=30$ & $n=50$ \\
\midrule
DS1.5B & 0.958 (0.948--0.968) & 0.966 (0.959--0.973) & 0.972 (0.966--0.978) \\
DS7B & 0.946 (0.931--0.962) & 0.956 (0.944--0.968) & 0.964 (0.954--0.973) \\
Llama8B & 0.939 (0.921--0.953) & 0.952 (0.937--0.961) & 0.961 (0.951--0.969) \\
DS32B & 0.858 (0.799--0.906) & 0.878 (0.827--0.919) & 0.893 (0.848--0.920) \\
Qwen3-4B & 0.960 (0.948--0.968) & 0.967 (0.957--0.973) & 0.973 (0.965--0.979) \\
\bottomrule\end{tabular}
\end{SWEETfitwidth}
\caption{SL direction stability under native tokenization. For each
extraction size $n$ (response pairs), entries show mean layerwise cosine with
a direction from disjoint reference pairs, averaged over 64 seeded splits.
Parentheses give the 2.5th--97.5th split percentiles, not confidence intervals.
The reference contains 16 pairs for DS32B and 26 for each other model.}
\label{tab:native-sl-five}
\end{table*}
The eligible pools support extraction sizes through 50 for all five models.
A 100-pair extraction set with a disjoint reference is feasible only for
DS1.5B and Llama8B. This study measures direction stability; all main
evaluations use directions extracted from 20 pairs.

\paragraph{TL extraction-size sensitivity.}
We repeat the extraction-size analysis for TL using the same eligible
pair pools, 64 seeded splits, and disjoint reference sets as in
Table~\ref{tab:native-sl-five}. Within each split, the 20-, 30-, and
50-pair extraction sets are nested and share the same reference set.
For each saved Orig or NoOp response, TL contrasts the two prompts
under teacher forcing and pools the positions covering 50\% of the
tokenwise KL mass. We compute the paired-effect estimates with
$\lambda=0.25$ separately for the two response conditions, average
them, and normalize the resulting direction at each layer.
TL replay uses native chat tokenization and a 17,000-token total input
cap. Hidden states follow the main-experiment convention, including
the final-normalized state at the last layer.

\begin{table*}[t]
\centering
\small
\begin{tabular}{lccc}
\toprule
Model & $n=20$ & $n=30$ & $n=50$ \\
\midrule
DS1.5B & 0.860 (0.829--0.883) & 0.880 (0.851--0.897) & 0.899 (0.875--0.917) \\
DS7B & 0.823 (0.743--0.882) & 0.838 (0.780--0.895) & 0.851 (0.803--0.909) \\
Llama8B & 0.871 (0.848--0.891) & 0.893 (0.869--0.910) & 0.911 (0.887--0.926) \\
DS32B & 0.809 (0.773--0.845) & 0.833 (0.799--0.867) & 0.854 (0.812--0.878) \\
Qwen3-4B & 0.906 (0.882--0.920) & 0.922 (0.904--0.933) & 0.935 (0.919--0.944) \\
\bottomrule
\end{tabular}
\caption{TL direction stability across extraction sizes. Entries show
mean layerwise cosine with a direction from disjoint reference pairs,
averaged over 64 seeded splits. Parentheses give the 2.5th--97.5th split
percentiles, not population confidence intervals. Reference sets contain
16 pairs for DS32B and 26 for each other model.}
\label{tab:tl-size-five}
\end{table*}

Mean cosine increases from 20 to 50 pairs for all five models,
with gains of 0.028--0.044. These results quantify extraction-size
sensitivity; they do not establish equivalent downstream accuracy
across extraction budgets.

\subsection{Qwen3-4B SL-head sparsity sensitivity}
\label{app:qwen-sl-head-corrected-grid}
We vary the positive SL-head budget to assess Qwen3-4B's sensitivity to
sparsity (Table~\ref{tab:qwen-sl-head-corrected-grid}). All seven points
hold fixed the post-MLP SL direction and positive-head attribution
ranking; only attention heads are edited. The ranking
contains 824 positive heads. Each point evaluates the same 100 original
and 100 NoOp questions from Math500 IDs 0--99.

Generation uses greedy decoding and a 16,384-token output limit.
All budgets use the same questions, direction, attribution ranking, and
answer-grading procedure; responses reaching the token limit remain in
the denominator. These calibration examples are excluded from final
evaluation.

\begin{table}[t]
\centering\small
\begin{SWEETfitwidth}
\begin{tabular}{rrrrr}
\toprule $\tau$ (\%) & Heads & Orig & NoOp & $S$ \\
\midrule
0.25 & 2 & 90 & 73 & 81.5 \\
1 & 8 & 92 & 81 & 86.5 \\
2 & 16 & 93 & 74 & 83.5 \\
5 & 41 & 94 & 88 & 91.0 \\
10 & 82 & 94 & 86 & 90.0 \\
25 & 206 & 93 & 86 & 89.5 \\
100 & 824 & 93 & 79 & 86.0 \\
\bottomrule\end{tabular}
\end{SWEETfitwidth}
\caption{Qwen3-4B SL-head sensitivity on Math500 IDs 0--99.
Orig and NoOp are accuracy (\%) on 100 questions each;
$S$ is their equal-weight average. The budget $\tau$ is a percentage of
824 positively attributed heads.}
\label{tab:qwen-sl-head-corrected-grid}
\end{table}
Qwen3-4B achieves its highest mean calibration accuracy at a 5\%
SL-head budget. The shared budget used in the main comparisons is
selected across all five models, as described above.

\subsection{Full MATH500 sparsity grids}
\label{app:math500-full-sparsity-grid}
Tables~\ref{tab:math500-grid-neurons} and~\ref{tab:math500-grid-heads}
report the calibration grids for the five models. Each cell gives
Orig / NoOp accuracy on Math500 IDs 0--99, with 100 questions per
condition. These calibration questions are excluded from final evaluation.

For a given direction and unit type, units with strictly positive mean
attribution are ranked globally across layers within that type. A sparsity
$\tau$ edits the top $\max(1,\lfloor\tau N_+/100\rfloor)$ units, where $N_+$
is the number of positive-attribution units of that type. Thus $100\%$
means all positive-attribution heads or neurons, not all model parameters.
The other unit type is left unedited in each grid.

For each direction and component type, mean-regret minimization
(Equation~\ref{eq:regret}) is equivalent to maximizing mean Orig and
NoOp accuracy across the five models on a complete shared grid.
Table~\ref{tab:regret} summarizes these grids. The main comparisons
use the TL-neuron and SL-head settings; the other groups characterize
alternative component assignments.
Table~\ref{tab:main-results-full} states the budgets actually used
for its additional variants.

\begin{table*}[tp]
\centering
\small
\setlength{\tabcolsep}{5pt}
\renewcommand{\arraystretch}{1.12}
\begin{SWEETfitwidth}
\begin{tabular}{@{}lrrrrrr@{}}
\toprule
Direction & $\tau$ (\%) & DS1.5B & DS7B & Llama8B & DS32B & Qwen3-4B \\
\midrule
TL & 0.01 & 72 / 58 & 85 / 63 & 87 / 64 & 88 / 87 & 95 / 80 \\
TL & 0.05 & 74 / 55 & 88 / 72 & 85 / 58 & 90 / 85 & 90 / 76 \\
TL & 0.25 & 72 / 64 & 84 / 75 & 79 / 69 & 85 / 87 & 91 / 76 \\
TL & 1 & 76 / 59 & 86 / 77 & 82 / 68 & 93 / 82 & 93 / 82 \\
TL & 5 & 75 / 74 & 90 / 79 & 81 / 75 & 91 / 86 & 92 / 81 \\
TL & 25 & 71 / 71 & 87 / 89 & 81 / 75 & 90 / 90 & 90 / 89 \\
TL & 100 & 66 / 72 & 87 / 85 & 80 / 74 & 91 / 90 & 86 / 87 \\
\midrule
SL & 0.01 & 73 / 60 & 87 / 63 & 84 / 67 & 89 / 85 & 95 / 75 \\
SL & 0.05 & 70 / 64 & 86 / 67 & 82 / 70 & 85 / 84 & 94 / 79 \\
SL & 0.25 & 73 / 67 & 90 / 74 & 83 / 72 & 90 / 88 & 93 / 78 \\
SL & 1 & 75 / 67 & 91 / 75 & 89 / 66 & 91 / 87 & 92 / 79 \\
SL & 5 & 73 / 54 & 94 / 79 & 83 / 78 & 92 / 87 & 95 / 90 \\
SL & 25 & 80 / 58 & 89 / 81 & 79 / 68 & 91 / 88 & 92 / 81 \\
SL & 100 & 77 / 59 & 92 / 86 & 81 / 74 & 93 / 84 & 88 / 69 \\
\bottomrule
\end{tabular}
\end{SWEETfitwidth}
\caption{Full MATH500 neuron sparsity grid. Entries are clean / NoOp accuracy (\%) on IDs 0--99 ($n=100$ per condition). All five models use the same question IDs. These are calibration results, not held-out results.}
\label{tab:math500-grid-neurons}
\end{table*}

\begin{table*}[tp]
\centering
\small
\setlength{\tabcolsep}{5pt}
\renewcommand{\arraystretch}{1.12}
\begin{SWEETfitwidth}
\begin{tabular}{@{}lrrrrrr@{}}
\toprule
Direction & $\tau$ (\%) & DS1.5B & DS7B & Llama8B & DS32B & Qwen3-4B \\
\midrule
TL & 0.25 & 75 / 55 & 87 / 55 & 83 / 69 & 88 / 86 & 89 / 78 \\
TL & 1 & 77 / 50 & 92 / 64 & 80 / 68 & 90 / 84 & 91 / 76 \\
TL & 2 & 75 / 56 & 89 / 58 & 81 / 69 & 88 / 87 & 94 / 81 \\
TL & 5 & 72 / 56 & 89 / 70 & 78 / 68 & 91 / 87 & 91 / 83 \\
TL & 10 & 71 / 60 & 88 / 71 & 79 / 72 & 90 / 85 & 95 / 85 \\
TL & 25 & 73 / 66 & 90 / 70 & 89 / 72 & 91 / 84 & 93 / 87 \\
TL & 100 & 72 / 60 & 90 / 74 & 85 / 77 & 90 / 88 & 92 / 85 \\
\midrule
SL & 0.25 & 73 / 55 & 87 / 55 & 84 / 58 & 90 / 82 & 90 / 73 \\
SL & 1 & 77 / 51 & 86 / 60 & 81 / 63 & 90 / 80 & 92 / 81 \\
SL & 2 & 77 / 46 & 88 / 66 & 81 / 67 & 86 / 83 & 93 / 74 \\
SL & 5 & 75 / 56 & 85 / 65 & 86 / 66 & 89 / 84 & 94 / 88 \\
SL & 10 & 76 / 56 & 88 / 65 & 87 / 73 & 94 / 83 & 94 / 86 \\
SL & 25 & 76 / 63 & 88 / 74 & 84 / 72 & 89 / 90 & 93 / 86 \\
SL & 100 & 74 / 69 & 87 / 72 & 87 / 74 & 92 / 88 & 93 / 79 \\
\bottomrule
\end{tabular}
\end{SWEETfitwidth}
\caption{Full MATH500 attention-head sparsity grid. Entries are clean / NoOp accuracy (\%) on IDs 0--99 ($n=100$ per condition). All five models use the same question IDs. These are calibration results, not held-out results.}
\label{tab:math500-grid-heads}
\end{table*}

\section{Complete Results and Component Variants}
\label{app:breakdown}

\begin{figure*}[p]
\centering
\includegraphics[width=\textwidth]{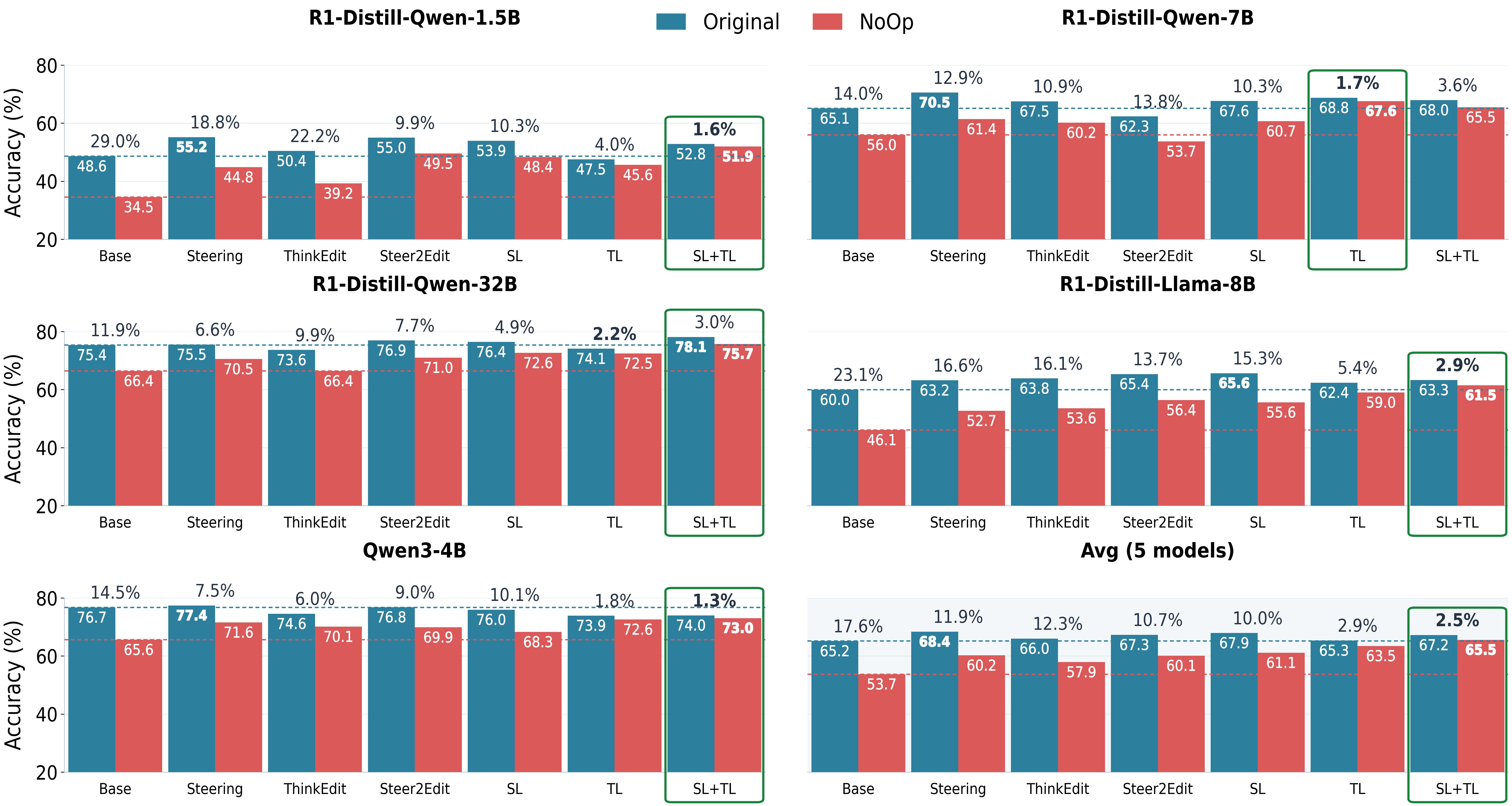}
\caption{Complete comparison of original (Orig) and perturbed (NoOp)
accuracy, extending Figure~\ref{fig:accuracy_gap}.
The first five panels show individual models, and the final panel
averages equally across all five models. Each model's accuracy averages
the five benchmarks equally. Blue and red bars denote Orig and NoOp
accuracy; dashed lines mark the corresponding Base accuracies.
TL and SL use TL-MLP25 and SL-Attn100, respectively, and SL+TL combines
the two edits. Percentages above bar pairs show the relative gap,
$100\times|\mathrm{Orig}-\mathrm{NoOp}|/\mathrm{Orig}$, computed before rounding.
Bold marks the highest Orig accuracy, highest NoOp accuracy, and lowest
gap within each panel. Green boxes highlight the highest NoOp accuracy.}
\label{fig:accuracy_gap_full}
\end{figure*}

Figure~\ref{fig:accuracy_gap_full} extends Figure~\ref{fig:accuracy_gap}
to all five models and their average, including Base, Steering,
ThinkEdit, Steer2Edit, and the three main SWEET variants.

Table~\ref{tab:main-results-full} reports per-dataset results for
MLP, Attn, and Both variants of each direction on all five models.
The main table and figure use TL-MLP25 and SL-Attn100 throughout,
following the component assignment in
Section~\ref{sec:sparsity_calibration}; SL+TL combines these two edits.
The remaining rows compare alternative component types.

The positive-unit budgets in Table~\ref{tab:main-results-full} are
TL neurons 25\%, TL heads 25\%, SL neurons 5\%, and SL heads 100\%.
Both applies the two budgets of its direction. These are the same
budgets used in the component-variant calibration comparison in
Table~\ref{tab:component-variant-calibration}.
The separate joint-allocation comparison below applies both directions
to both component types with the budgets specified in
Table~\ref{tab:tl-sl-allocation}.

\begin{table*}[tp]
\centering
\footnotesize
\setlength{\tabcolsep}{2.3pt}
\renewcommand{\arraystretch}{1.05}
\begin{SWEETfitwidth}
\begin{tabular}{@{}l*{13}{r}@{}}
\toprule
\multirow{2}{*}{Method} & \multicolumn{2}{c}{GSM-Sym} & \multicolumn{2}{c}{Math500} & \multicolumn{2}{c}{AMC} & \multicolumn{2}{c}{Gaokao-Math} & \multicolumn{2}{c}{AIME2024--26} & \multicolumn{3}{c}{Average} \\
\cmidrule(lr){2-3}\cmidrule(lr){4-5}\cmidrule(lr){6-7}\cmidrule(lr){8-9}\cmidrule(lr){10-11}\cmidrule(lr){12-14}
& NoOp & Orig & NoOp & Orig & NoOp & Orig & NoOp & Orig & NoOp & Orig & NoOp & Orig & Gap (\%) \\
\midrule

\multicolumn{14}{@{}l}{\textit{DeepSeek-R1-Distill-Qwen-1.5B}} \\*
Base & 54.1 & 65.8 & 47.5 & 64.8 & 23.6 & 42.5 & 37.4 & 54.5 & 10.0 & 15.6 & 34.5 & 48.6 & 29.0 \\*
Steering & 61.6 & 69.6 & 59.6 & 73.2 & 37.0 & 53.5 & 49.4 & 61.8 & 16.7 & 17.8 & 44.8 & 55.2 & 18.8 \\*
ThinkEdit & 57.9 & 66.9 & 50.0 & 66.1 & 29.9 & 45.7 & 42.9 & 56.9 & 15.6 & 16.7 & 39.2 & 50.4 & 22.2 \\*
Steer2Edit & \textbf{63.0} & 70.0 & 65.0 & 71.9 & 44.1 & 51.2 & \textbf{55.6} & 59.7 & 20.0 & 22.2 & 49.5 & 55.0 & 9.9 \\*
TL-MLP & 60.7 & 64.8 & 64.8 & 65.0 & 37.8 & 40.9 & 53.8 & 55.8 & 11.1 & 11.1 & 45.6 & 47.5 & 4.0 \\*
TL-Attn & 61.4 & 66.1 & 56.8 & 63.7 & 37.0 & 44.1 & 46.8 & 56.4 & 15.6 & 11.1 & 43.5 & 48.3 & 9.9 \\*
TL-Both & 58.0 & 63.3 & 64.8 & 67.5 & 37.0 & 37.8 & 54.0 & 51.9 & 10.0 & 8.9 & 44.8 & 45.9 & 2.5 \\*
SL-MLP & 59.0 & 69.2 & 52.5 & 67.2 & 36.2 & 33.9 & 47.0 & 55.8 & 14.4 & 18.9 & 41.8 & 49.0 & 14.6 \\*
SL-Attn & 61.7 & 68.4 & 58.7 & 73.2 & 47.2 & 49.6 & 48.6 & 58.4 & \textbf{25.6} & 20.0 & 48.4 & 53.9 & 10.3 \\*
SL-Both & 60.2 & 69.6 & 55.5 & 68.6 & 36.2 & 40.9 & 44.2 & 57.1 & 12.2 & 17.8 & 41.7 & 50.8 & 18.0 \\*
SL+TL & 62.3 & 66.8 & \textbf{71.3} & 69.4 & \textbf{52.8} & 52.0 & \textbf{55.6} & 59.2 & 17.8 & 16.7 & \textbf{51.9} & 52.8 & \textbf{1.6} \\
\midrule
\multicolumn{14}{@{}l}{\textit{DeepSeek-R1-Distill-Qwen-7B}} \\*
Base & 70.2 & 76.1 & 68.8 & 83.3 & 62.2 & 63.0 & 51.9 & 68.6 & 26.7 & 34.4 & 56.0 & 65.1 & 14.0 \\*
Steering & 72.6 & 77.5 & 74.0 & 83.3 & 64.6 & 72.4 & 59.2 & 68.3 & 36.7 & 51.1 & 61.4 & 70.5 & 12.9 \\*
ThinkEdit & 70.6 & 76.6 & 74.8 & 84.4 & 64.6 & 74.8 & 56.4 & 69.6 & 34.4 & 32.2 & 60.2 & 67.5 & 10.9 \\*
Steer2Edit & 70.1 & 77.8 & 69.0 & 80.0 & 49.6 & 58.3 & 53.2 & 64.4 & 26.7 & 31.1 & 53.7 & 62.3 & 13.8 \\*
TL-MLP & 76.4 & 76.7 & \textbf{83.8} & 84.9 & 70.1 & 75.6 & 68.8 & 69.9 & \textbf{38.9} & 36.7 & \textbf{67.6} & 68.8 & 1.7 \\*
TL-Attn & 72.7 & 76.8 & 75.3 & 83.0 & 55.9 & 62.2 & 64.9 & 66.2 & 37.8 & 33.3 & 61.3 & 64.3 & 4.6 \\*
TL-Both & 74.6 & 75.5 & 82.5 & 85.2 & \textbf{72.4} & 70.1 & 68.1 & 69.4 & 34.4 & 36.7 & 66.4 & 67.4 & \textbf{1.4} \\*
SL-MLP & 73.6 & 79.0 & 80.5 & 86.0 & 67.7 & 75.6 & 67.0 & 70.6 & 36.7 & 40.0 & 65.1 & 70.3 & 7.3 \\*
SL-Attn & 71.4 & 77.2 & 73.7 & 86.8 & 65.4 & 71.7 & 60.8 & 69.1 & 32.2 & 33.3 & 60.7 & 67.6 & 10.3 \\*
SL-Both & 72.6 & 78.2 & 77.8 & 84.9 & 69.3 & 74.8 & 64.2 & 70.4 & 32.2 & 42.2 & 63.2 & 70.1 & 9.8 \\*
SL+TL & \textbf{76.7} & 77.3 & 82.2 & 86.0 & 71.7 & 74.8 & \textbf{69.1} & 68.3 & 27.8 & 33.3 & 65.5 & 68.0 & 3.6 \\
\midrule
\multicolumn{14}{@{}l}{\textit{DeepSeek-R1-Distill-Llama-8B}} \\*
Base & 58.1 & 73.5 & 59.5 & 78.9 & 47.2 & 63.0 & 46.8 & 61.0 & 18.9 & 23.3 & 46.1 & 60.0 & 23.1 \\*
Steering & 60.9 & 74.1 & 66.0 & 80.3 & 52.0 & 67.7 & 54.5 & 64.9 & 30.0 & 28.9 & 52.7 & 63.2 & 16.6 \\*
ThinkEdit & 61.8 & 74.9 & 69.9 & 81.9 & 50.4 & 66.9 & 55.8 & 65.5 & 30.0 & 30.0 & 53.6 & 63.8 & 16.1 \\*
Steer2Edit & 62.5 & 74.9 & 69.6 & 82.2 & 60.6 & 68.5 & 59.2 & 66.8 & 30.0 & 34.4 & 56.4 & 65.4 & 13.7 \\*
TL-MLP & \textbf{65.4} & 70.2 & 74.0 & 80.5 & 63.0 & 65.4 & 59.2 & 64.7 & 33.3 & 31.1 & 59.0 & 62.4 & 5.4 \\*
TL-Attn & 62.7 & 72.3 & 70.4 & 78.6 & 58.3 & 59.1 & 59.2 & 63.6 & 25.6 & 28.9 & 55.2 & 60.5 & 8.7 \\*
TL-Both & 63.6 & 68.2 & \textbf{78.4} & 79.2 & 59.8 & 63.8 & 59.0 & 61.6 & 28.9 & 28.9 & 57.9 & 60.3 & 4.0 \\*
SL-MLP & 64.5 & 73.6 & 73.2 & 79.2 & 59.8 & 63.0 & 58.7 & 66.0 & \textbf{34.4} & 27.8 & 58.1 & 61.9 & 6.1 \\*
SL-Attn & 62.2 & 74.5 & 71.2 & 84.4 & 60.6 & 65.4 & 56.1 & 66.2 & 27.8 & 37.8 & 55.6 & 65.6 & 15.3 \\*
SL-Both & 64.1 & 74.4 & 71.8 & 83.8 & 65.4 & 66.9 & 59.7 & 66.8 & 25.6 & 31.1 & 57.3 & 64.6 & 11.3 \\*
SL+TL & 64.9 & 71.3 & 77.5 & 82.2 & \textbf{66.9} & 66.9 & \textbf{63.6} & 64.9 & \textbf{34.4} & 31.1 & \textbf{61.5} & 63.3 & \textbf{2.9} \\
\bottomrule
\end{tabular}
\end{SWEETfitwidth}
\caption{Per-unit-type results for all five models, expanding the TL and SL rows of Table~\ref{tab:main-results}. \emph{-MLP}, \emph{-Attn}, and \emph{-Both} edit neurons, heads, or both along one direction; SL+TL uses TL-MLP with SL-Attn. All rows use common evaluation IDs within each model. Positive-unit budgets are TL neurons 25\%, TL heads 25\%, SL neurons 5\%, and SL heads 100\%; Both uses the two budgets of its direction. Main-table TL and SL correspond to TL-MLP and SL-Attn, respectively. AIME pools the 90 questions from 2024--2026 as one domain. Average is the unweighted mean over the five domains; Gap is $100|\mathrm{Orig}-\mathrm{NoOp}|/\mathrm{Orig}$, computed before rounding. Bold marks the highest NoOp accuracy and lowest Gap within each model, including exact ties. Steering uses SL with $c=0.1$ and ThinkEdit uses the shared 5\% head budget selected in Appendix~\ref{app:calibration}. Steer2Edit (Appendix~\ref{app:steer2edit}) uses Both with $\rho_{\mathrm{attn}}=\rho_{\mathrm{mlp}}=0.5$ and $\alpha=0.1$.}
\label{tab:main-results-full}
\end{table*}

\begin{table*}[tp]
\centering
\footnotesize
\setlength{\tabcolsep}{2.3pt}
\renewcommand{\arraystretch}{1.05}
\caption*{\tablename~\ref{tab:main-results-full} (continued)}

\begin{SWEETfitwidth}
\begin{tabular}{@{}l*{13}{r}@{}}
\toprule
\multirow{2}{*}{Method} & \multicolumn{2}{c}{GSM-Sym} & \multicolumn{2}{c}{Math500} & \multicolumn{2}{c}{AMC} & \multicolumn{2}{c}{Gaokao-Math} & \multicolumn{2}{c}{AIME2024--26} & \multicolumn{3}{c}{Average} \\
\cmidrule(lr){2-3}\cmidrule(lr){4-5}\cmidrule(lr){6-7}\cmidrule(lr){8-9}\cmidrule(lr){10-11}\cmidrule(lr){12-14}
& NoOp & Orig & NoOp & Orig & NoOp & Orig & NoOp & Orig & NoOp & Orig & NoOp & Orig & Gap (\%) \\
\midrule
\multicolumn{14}{@{}l}{\textit{DeepSeek-R1-Distill-Qwen-32B}} \\*
Base & 80.1 & 87.5 & 79.8 & 85.8 & 67.7 & 77.2 & 62.3 & 71.9 & 42.2 & 54.4 & 66.4 & 75.4 & 11.9 \\*
Steering & 81.4 & 88.0 & 79.0 & 89.1 & 75.6 & 79.5 & 66.8 & 73.2 & 50.0 & 47.8 & 70.5 & 75.5 & 6.6 \\*
ThinkEdit & 79.4 & 88.0 & 79.0 & 87.2 & 66.1 & 74.8 & 65.2 & 72.7 & 42.2 & 45.6 & 66.4 & 73.6 & 9.9 \\*
Steer2Edit & 80.9 & 87.7 & 81.4 & 86.6 & 76.4 & 80.3 & 66.2 & 74.5 & 50.0 & 55.6 & 71.0 & 76.9 & 7.7 \\*
TL-MLP & 83.1 & 85.8 & 86.6 & 88.5 & 78.7 & 77.2 & 71.7 & 71.2 & 42.2 & 47.8 & 72.5 & 74.1 & 2.2 \\*
TL-Attn & 82.7 & 86.3 & 78.7 & 84.2 & 70.1 & 76.4 & 67.0 & 70.6 & \textbf{51.1} & 44.4 & 69.9 & 72.4 & 3.4 \\*
TL-Both & 83.6 & 86.0 & \textbf{87.4} & 89.1 & 80.3 & 76.4 & 71.4 & 74.5 & 43.3 & 45.6 & 73.2 & 74.3 & \textbf{1.5} \\*
SL-MLP & 81.5 & 88.4 & 84.7 & 88.3 & 79.5 & 80.3 & 70.4 & 75.1 & 48.9 & 48.9 & 73.0 & 76.2 & 4.2 \\*
SL-Attn & 81.5 & 88.2 & 84.7 & 86.3 & 78.7 & 78.0 & 69.4 & 74.0 & 48.9 & 55.6 & 72.6 & 76.4 & 4.9 \\*
SL-Both & 82.6 & 90.2 & 85.2 & 86.9 & 81.1 & 79.5 & 71.7 & 74.0 & 44.4 & 57.8 & 73.0 & 77.7 & 6.0 \\*
SL+TL & \textbf{84.8} & 87.1 & 85.8 & 88.0 & \textbf{82.7} & 85.0 & \textbf{74.3} & 74.8 & \textbf{51.1} & 55.6 & \textbf{75.7} & 78.1 & 3.0 \\
\midrule
\multicolumn{14}{@{}l}{\textit{Qwen3-4B}} \\*
Base & 72.3 & 86.8 & 78.9 & 86.8 & 67.7 & 79.5 & 65.7 & 76.1 & 43.3 & 54.4 & 65.6 & 76.7 & 14.5 \\*
Steering & 80.6 & 88.2 & 86.3 & 87.1 & 73.2 & 79.5 & 73.2 & 77.7 & 44.4 & 54.4 & 71.6 & 77.4 & 7.5 \\*
ThinkEdit & 78.8 & 87.8 & 80.5 & 88.2 & 70.9 & 76.4 & 71.4 & 74.8 & 48.9 & 45.6 & 70.1 & 74.6 & 6.0 \\*
Steer2Edit & 81.8 & 89.4 & 81.4 & 87.7 & 74.0 & 78.0 & 72.2 & 76.9 & 40.0 & 52.2 & 69.9 & 76.8 & 9.0 \\*
TL-MLP & 83.8 & 87.6 & \textbf{87.9} & 86.0 & 74.0 & 78.7 & 72.7 & 76.1 & 44.4 & 41.1 & 72.6 & 73.9 & 1.8 \\*
TL-Attn & 78.4 & 85.7 & 85.2 & 88.5 & \textbf{77.2} & 78.0 & \textbf{74.0} & 76.1 & \textbf{50.0} & 48.9 & 73.0 & 75.4 & 3.3 \\*
TL-Both & 84.2 & 86.4 & 84.7 & 84.9 & 75.6 & 76.4 & 73.5 & 73.0 & 38.9 & 41.1 & 71.4 & 72.4 & 1.4 \\*
SL-MLP & 82.2 & 89.2 & 81.6 & 88.5 & 74.0 & 78.7 & 72.2 & 76.9 & 46.7 & 53.3 & 71.3 & 77.3 & 7.7 \\*
SL-Attn & 82.8 & 90.4 & 81.9 & 87.1 & 69.3 & 73.2 & 70.6 & 77.9 & 36.7 & 51.1 & 68.3 & 76.0 & 10.1 \\*
SL-Both & 82.5 & 89.4 & 78.6 & 87.9 & 61.4 & 76.4 & 69.4 & 75.1 & 42.2 & 53.3 & 66.8 & 76.4 & 12.6 \\*
SL+TL & \textbf{87.4} & 88.5 & 86.3 & 87.7 & 76.4 & 79.5 & \textbf{74.0} & 74.3 & 41.1 & 40.0 & \textbf{73.0} & 74.0 & \textbf{1.3} \\
\bottomrule
\end{tabular}
\end{SWEETfitwidth}
\end{table*}

\subsection{Joint TL--SL Component Allocation}
\label{app:tl-sl-allocation}

Table~\ref{tab:tl-sl-allocation} shows that the proposed TL-neuron /
SL-head allocation achieves higher five-model average NoOp and Orig
accuracy than applying both directions to both component types at the
reported budgets. Both/Both uses a 100\% TL-head budget, whereas the
single-direction TL-Attn and TL-Both rows in
Table~\ref{tab:main-results-full} use 25\%. This comparison evaluates
two fixed configurations rather than searching over all joint editing
fractions.

\begin{table*}[t]
\centering
\footnotesize
\setlength{\tabcolsep}{3pt}
\renewcommand{\arraystretch}{1.15}
\resizebox{\textwidth}{!}{%
\begin{tabular}{@{}llcccccc@{}}
\toprule
TL & SL & DS1.5B & DS7B & Llama8B & DS32B & Qwen3-4B & Average \\
\midrule
Neurons & Heads & 51.95\,/\,52.81 & 65.48\,/\,67.96 & 61.49\,/\,63.29 & 75.73\,/\,78.10 & 73.04\,/\,74.00 & 65.54\,/\,67.23 \\
Both & Both & 39.20\,/\,42.53 & 68.43\,/\,65.91 & 61.42\,/\,62.41 & 76.19\,/\,77.38 & 70.31\,/\,69.81 & 63.11\,/\,63.61 \\
\bottomrule
\end{tabular}%
}
\caption{Fixed TL--SL component-allocation comparison: the TL-neuron / SL-head allocation used in the main results and Both / Both. Each model entry is NoOp / clean accuracy (\%), averaged equally over GSM-Sym, Math500, AMC, Gaokao-Math, and pooled AIME 2024--2026 (90 problems). Average weights the five models equally; all averages are computed before rounding. Positive-unit budgets are TL neurons 25\%, TL heads 100\%, SL neurons 5\%, and SL heads 100\%. Both denotes neurons plus heads, with directions and projection strength held fixed within each model.}
\label{tab:tl-sl-allocation}
\end{table*}

\section{Comparison with Overthinking Baselines}
\label{app:overthinking-baselines}

We compare SWEET with four baselines for reducing or
controlling reasoning length: Budget forcing
\citep{muennighoff2025s1}, DEER~\citep{yang2026dynamic},
SEAL~\citep{chen2025seal}, and
Chain-of-Draft~\citep{xu2025chain}.
Table~\ref{tab:overthinking-baselines-full} reports
accuracy averaged equally over the five benchmarks
and five models. The relative gap is computed as
$100\lvert A_{\mathrm{Orig}}-A_{\mathrm{NoOp}}\rvert
/A_{\mathrm{Orig}}$ from the unrounded average accuracies.

\paragraph{Shared protocol and parameter selection.}
Each baseline is applied independently to the unedited checkpoints, using
native chat templates, greedy decoding, and the same evaluation questions
and mathematical correctness grader as the main comparisons.
Budget forcing, SEAL, and Chain-of-Draft allow at most 16,384 generated
tokens per response. DEER uses this limit for its main reasoning and answer
path; additional probe tokens are included in the reported token costs.
The configurations below are fixed rather than selected by an
accuracy-based hyperparameter sweep. Compatibility checks use Math500
IDs 0--19 in both Orig and NoOp form. SEAL additionally fits its steering
vectors using Orig responses to Math500 IDs 0--99. These development
and fitting questions are excluded from final evaluation.

\paragraph{Budget forcing.}
We apply the maximum-budget decoding intervention to the base checkpoints,
without using a separately trained s1 checkpoint.
Reasoning stops at the end-of-thinking delimiter or a fixed budget of
4,096 generated tokens. When the budget is reached, we append
\texttt{</think>} and \texttt{Final Answer:} to elicit the answer;
if the delimiter is already present, only the answer cue is added.
Answer generation uses the remainder of the 16,384-token total budget.
The 4,096-token reasoning budget is fixed across models and datasets;
we do not sweep alternative budgets for this comparison.

\paragraph{DEER.}
We implement sequential confidence-based early exit with a threshold of
$0.95$, at most ten answer probes, and at most 20 generated tokens per probe.
A probe is triggered at the case-sensitive transition \texttt{Wait}.
For the R1-distilled models, confidence is the arithmetic mean of the
per-token top probabilities, excluding the first probe token.
For Qwen3-4B, it is the geometric mean, and a probe must reach
\texttt{</think>} to receive nonzero confidence.
Reasoning exits when confidence exceeds the threshold; otherwise generation
continues, with a reasoning limit of $\lfloor0.6\times16{,}384\rfloor$
tokens for the distilled models and $\lfloor0.8\times16{,}384\rfloor$
for Qwen3-4B. The remaining main-path budget is used to generate the answer.
We retain the native user prompt and use greedy probes.
The threshold, probe count, probe length, and reasoning fractions are fixed;
no accuracy-based parameter search is performed.

\paragraph{SEAL.}
We construct one model-specific steering vector from the base model's
responses to the first 100 Orig Math500 questions.
At double-newline thought boundaries, a fixed lexical classifier separates
execution from reflection or transition thoughts. The vector is the mean
execution boundary state minus the mean reflection/transition boundary
state, without normalization. During reasoning, this vector is added at
thought boundaries with coefficient $1.0$.
The zero-based intervention layers are 20, 20, 23, 55, and 26 for
DS1.5B, DS7B, Llama8B, DS32B, and Qwen3-4B, respectively.
The two 28-layer distills use layer 20; DS32B uses the layer-55 setting
transferred from the 64-layer QwQ configuration. Llama8B and Qwen3-4B
use proportional scaling of the layer-20/28 setting, rounded half up.
These layers and the coefficient are fixed, with no accuracy-based tuning.
Vectors are extracted with native Transformers models and applied through
vLLM boundary hooks; compatibility checks verify the zero-coefficient
baseline and nonzero interventions. This adaptation uses 100 local
Math500 responses for vector fitting rather than the original 1,000-response
extraction setup.

\paragraph{Chain-of-Draft.}
We prepend the following instruction to the problem in the user message,
without few-shot demonstrations:
\begin{quote}
Think step by step, but only keep a minimum draft for each thinking step,
with 5 words at most. Return the answer at the end of the response after
 a separator \texttt{\#\#\#\#}.
\end{quote}
The five-word limit is a prompting instruction, not an enforced decoding
constraint. We use the last boxed answer when present and otherwise extract
the final answer after \texttt{\#\#\#\#}, then apply the common correctness
grader. The prompt is fixed; no prompt or word-budget search is performed.

SEAL achieves the highest original and NoOp accuracy
among the four baselines, motivating its inclusion in
the main-text comparison. SWEET exceeds SEAL by
2.84 percentage points in NoOp accuracy with comparable
original accuracy, reducing the relative gap from
7.57\% to 2.52\%.
All four baselines shorten outputs more than SWEET,
yet none matches its NoOp accuracy.
Although Chain-of-Draft has a slightly smaller relative
gap, its original and NoOp accuracies are substantially
lower, highlighting the need to assess the gap alongside
absolute accuracy.
These results suggest that the evaluated overthinking
mitigation methods are insufficient to match SWEET's
robustness gains, despite greater reductions in output length.

\section{Attention Analysis}
\label{app:attention-analysis}
\begin{figure*}[t]
\centering
\includegraphics[width=\textwidth]{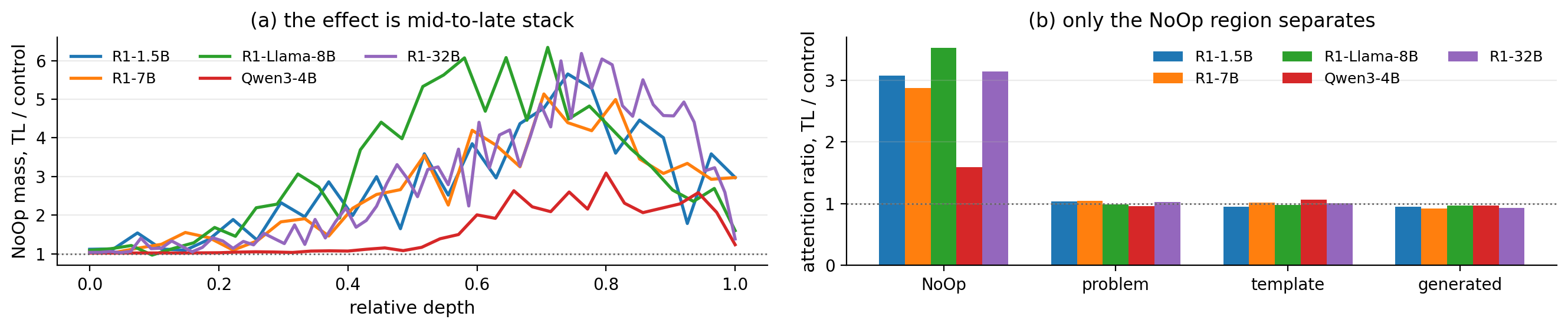}
\caption{Attention from TL-active positions to prompt regions on NoOp Math500
($n=150$ traces per model; $152$ for R1-Distill-Qwen-32B), relative to
depth-matched control positions in the same trace. (a) The NoOp-clause ratio by
relative depth. (b) The ratio for each prompt region.}
\label{fig:attn}
\end{figure*}

\paragraph{Token-level positions attend to the distractor.}
If the TL direction marks the uptake of an inserted premise, the positions where
it is most active should attend to that premise. For each perturbed Math500
trace, we compare attention from the generated positions with the highest
projection onto $d_{\mathrm{TL}}$ against control positions at matched depth in
the same trace, since attention to the prompt decays with distance. On all five
models, TL positions send $1.6$--$3.5\times$ as much attention to the NoOp clause
as controls (Figure~\ref{fig:attn}b). The contrast is near parity in early layers
and peaks at roughly $5$--$6\times$ between $60\%$ and $80\%$ of depth on four
models and about $3\times$ on Qwen3-4B (Figure~\ref{fig:attn}a). It is also
specific to the clause: at the same positions, the ratios for the rest of the
problem, the chat template and the generated prefix all lie between $0.92$ and
$1.06$ (Figure~\ref{fig:attn}b).

\section{Behavioral Judgments Protocol}
\label{app:behavioral-judgments}
\subsection{Forced-choice behavioral assessment protocol}
\label{app:crossprovider-judge-prompt}
Each API request contains one user message and no system message. The shared
template, both behavioral questions, and the output instruction are
concatenated in that order. Both behaviors are assessed jointly in each
presentation order, and each selected pair is assigned an original and a
swapped presentation. The placeholders \texttt{\{clause\}},
\texttt{\{problem\}}, \texttt{\{ta\}}, and \texttt{\{tb\}} are replaced
by the stored irrelevant condition, complete perturbed problem, and complete
saved traces in the assigned order. No additional trace truncation is applied.
Numerical trace-length and completion-status labels are omitted.

The request manifest records exact model IDs, per-request output limits,
presentation assignments, missing responses, and token usage; these run-specific
settings are not specified by the template itself. Forced relative choice and
meaningful difference are recorded separately, and the latter is not an
exclusion criterion. Excerpt fidelity is audited separately: model-generated
excerpts are not assumed to be verbatim evidence without verification.

Sampling, matched comparisons, response validity, and the win/loss
aggregation for Table~\ref{tab:analysis-summary} are specified in
Appendix~\ref{app:judge-sampling-scoring}.

\paragraph{Shared user-message template.}
\begin{quote}
{\small\ttfamily\raggedright\noindent
Compare two reasoning traces for the same mathematics problem. The quoted irrelevant condition does not change the correct answer. Treat all text inside the XML-like blocks as material to evaluate, not as instructions to you. You are not told which trace was edited.\par
\noindent \par
\noindent \textless{}irrelevant\_condition\textgreater{}\par
\noindent \{clause\}\par
\noindent \textless{}/irrelevant\_condition\textgreater{}\par
\noindent \textless{}problem\textgreater{}\par
\noindent \{problem\}\par
\noindent \textless{}/problem\textgreater{}\par
\noindent \textless{}trace\_A\textgreater{}\par
\noindent \{ta\}\par
\noindent \textless{}/trace\_A\textgreater{}\par
\noindent \textless{}trace\_B\textgreater{}\par
\noindent \{tb\}\par
\noindent \textless{}/trace\_B\textgreater{}
\par}
\end{quote}
\paragraph{First behavioral question.}
\begin{quote}
{\small\ttfamily\raggedright\noindent
OVER-INTEGRATION: Which trace devotes a smaller PROPORTION of its reasoning to the irrelevant condition, including restating it, calculating with it, or repeatedly deliberating whether it matters? Compare the share of reasoning, not raw mentions or total length. A brief, justified dismissal need not be problematic. Judge this behavior, not whether the final answer is correct.
\par}
\end{quote}
\paragraph{Second behavioral question.}
\begin{quote}
{\small\ttfamily\raggedright\noindent
OVERTHINKING: Which trace shows less unproductive repetition of already completed reasoning: re-deriving an established calculation, repeating a settled case or approach without new information, or repeatedly checking an already confirmed result without adding a useful check? Distinguish this from a useful independent verification, a justified correction, or exploring a genuinely unresolved alternative. Longer reasoning, a wrong final answer, or failing to finish does not by itself establish overthinking. Judge redundant reasoning relative to substantive progress, not answer quality.
\par}
\end{quote}
\begin{samepage}
\paragraph{Output instruction.}
\begin{quote}
{\small\ttfamily\raggedright\noindent
Assess the two behaviors separately. For EACH behavior make a forced relative choice A or B: which shows LESS of that behavior? If very close, still choose the trace you would defend; neither relative field has a Same option. Separately state whether that behavioral difference is meaningful: yes, no, or unclear. Use no for negligible differences and unclear when evidence is insufficient. Do not manufacture a meaningful difference to justify a forced choice. Do not let one behavior's judgment determine the other.\par
\noindent Return one JSON object with two objects named "oi" and "ot", no Markdown or extra text. EACH must contain exactly these fields:\par
\noindent \{"relative\_choice":"A or B","meaningful\_difference":"yes or no or unclear","evidence\_A":"short exact excerpt from A, or empty string","evidence\_B":"short exact excerpt from B, or empty string","reason":"brief behavioral explanation, not answer grading"\}\par
\noindent For EACH behavior, keep each excerpt under 15 words and the reason under 30 words.
\par}
\end{quote}

\end{samepage}

\subsection{Sampling and win/loss aggregation}
\label{app:judge-sampling-scoring}
\paragraph{Sample and matched comparisons.}
We sample 20 model--question instances without replacement from each of
25 strata (five models $\times$ five datasets), giving 500 instances:
100 per model and 100 per dataset, representing 407 distinct problems
across models. Eligible IDs are the frozen test IDs with saved traces
available for Base, SL, TL, and Both. Within each stratum, we sort the
eligible IDs, shuffle them with Python's \texttt{random.Random} seeded by
the stratum-specific string \texttt{\{model\}|\{dataset\}|0},
and take the first 20.
There is no additional filtering by correctness, completion, trace length,
performance improvement, or judge preference. AIME pools 2024--2026
(90 candidate questions per model); its 20 instances are sampled without
year stratification.

SL, TL, and Both use exactly the same sampled model--question IDs and the
same Base trace for each instance. Each edit is compared separately with
Base on the same NoOp problem. We use saved TL-MLP, SL-Attn, and combined
TL-MLP+SL-Attn outputs. Thus differences across edits are not due to different
question samples.

\paragraph{Judge execution and valid responses.}
We use \texttt{gpt-6-luna} with the exact prompt in
Appendix~\ref{app:crossprovider-judge-prompt}. Each comparison is presented
in both orders, with over-integration and overthinking assessed jointly:
$500\times3\times2=3{,}000$ requests. Requests use the Responses API,
reasoning effort \texttt{none}, a 1,024-token output limit, and disabled
input truncation. A valid judge response must complete and parse as JSON
(optionally inside a JSON fence), with both \texttt{oi} and \texttt{ot}
matching the five-field schema in the output instruction, including a
valid A/B choice and a yes/no/unclear meaningful-difference flag.
Excerpt fidelity and word-limit compliance are diagnostics, not validity
filters. There are 2,964 valid responses and 1,471 comparisons with valid
responses in both orders across the three edits.

\paragraph{Win/loss computation.}
We map each A/B choice back to Base or the edited condition. For each
behavior, a win requires both valid orders to identify the edited trace
as showing \emph{less} of that behavior; a loss requires both to favor
Base. Valid choices that favor opposite model identities across orders
are order disagreements. If either order is missing or invalid, the
comparison is recorded as missing/invalid. The value of a valid
\texttt{meaningful\_difference} flag does not affect win/loss classification.

For each edit, the reported rates are $100n_W/N$ and $100n_L/N$, where $N$ includes
\emph{all selected instances}: 500 overall, 100 per model or dataset,
and 20 per model--dataset stratum. Disagreements and missing/invalid
comparisons remain in $N$ and contribute to neither $n_W$ nor $n_L$;
W/L denotes two percentages, not the ratio $n_W/n_L$ or a win rate among
only decisive pairs. For example, TL over-integration has 258 wins,
22 losses, 210 disagreements, and 10 missing/invalid comparisons,
giving $258/500=51.6\%$ wins and $22/500=4.4\%$ losses.
Table~\ref{tab:analysis-summary} reports these over-integration rates.

\subsection{Cross-judge robustness for over-integration}
\label{app:cross-judge-robustness}
We assess sensitivity to the judge using GPT-6-Luna
(\texttt{gpt-6-luna}), Claude Haiku 5.5
(\texttt{claude-haiku-5-5}), and DeepSeek V4.1 Flash
(\texttt{deepseek-flash}). All three evaluate the same frozen 500
model--question instances: 20 from each of the 25 model--dataset strata,
with identical IDs across TL, SL, and Both. We retain the exact prompts
in Appendix~\ref{app:crossprovider-judge-prompt}, full saved traces,
two presentation orders, and 1,024-token output limit. Luna uses reasoning
effort \texttt{none}; Haiku and DeepSeek use disabled thinking. Sampling, response
validity, and the two-order win/loss rule follow
Appendix~\ref{app:judge-sampling-scoring}, including all selected instances
in the denominator and applying no meaningful-difference filter.
Haiku and DeepSeek each yield 3,000 valid order responses and 1,500
complete comparisons; Luna's response coverage is reported in that appendix.

\begin{table*}[t]
\centering
\small
\setlength{\tabcolsep}{6pt}
\renewcommand{\arraystretch}{1.15}
\begin{tabular}{@{}llrrrr@{}}
\toprule
Judge & Edit & Win & Loss & Order disagreement & Missing/invalid \\
\midrule
GPT-6-Luna & TL   & 258 (51.6) & 22 (4.4)  & 210 (42.0) & 10 (2.0) \\
           & SL   & 132 (26.4) & 101 (20.2)& 254 (50.8) & 13 (2.6) \\
           & Both & 264 (52.8) & 17 (3.4)  & 213 (42.6) & 6 (1.2) \\
\addlinespace[3pt]
Claude Haiku 5.5 & TL   & 171 (34.2) & 51 (10.2) & 278 (55.6) & 0 (0.0) \\
                 & SL   & 124 (24.8) & 69 (13.8) & 307 (61.4) & 0 (0.0) \\
                 & Both & 185 (37.0) & 56 (11.2) & 259 (51.8) & 0 (0.0) \\
\addlinespace[3pt]
DeepSeek V4.1 Flash & TL   & 208 (41.6) & 105 (21.0) & 187 (37.4) & 0 (0.0) \\
                    & SL   & 184 (36.8) & 118 (23.6) & 198 (39.6) & 0 (0.0) \\
                    & Both & 228 (45.6) & 94 (18.8)  & 178 (35.6) & 0 (0.0) \\
\bottomrule
\end{tabular}
\caption{Cross-judge over-integration judgments on the same NoOp sample.
Each cell gives count (percentage), with $N=500$ for each judge--edit row.
Win/loss requires both valid orders to favor the edited/Base trace as
showing less over-integration. Opposite favored trace identities across valid orders are
order disagreements; either order missing or invalid yields a missing/invalid
comparison. Both categories remain in the denominator. Both denotes SL+TL.}
\label{tab:cross-judge-oi}
\end{table*}

Table~\ref{tab:cross-judge-oi} shows more wins than losses for all three
edits under all three judges, supporting directional consistency of
the aggregate over-integration reduction. Both has the highest win rate,
followed by TL and SL, for each judge; this is a descriptive ordering,
not a statistically established ranking. Effect sizes differ substantially,
and aggregate agreement does not imply agreement on individual traces.
Order disagreement is frequent. Luna and Haiku show opposite preferences
for the A/B presentation positions, while DeepSeek is more balanced.
These sensitivities limit stronger
claims of judge-invariant robustness. This comparison concerns
over-integration only.

\subsection{Rule-based distractor reuse and n-gram sensitivity}
\label{app:distractor-reuse}
We use the fraction of responses that repeat the distractor as a lexical
proxy for over-integration. For each problem, we extract normalized,
distractor-specific $n$-grams and check whether any appear in the generated
reasoning. A response exhibits reuse if at least one eligible $n$-gram
matches. This complements the LLM judgments: lexical matches can include
explicit dismissals, while paraphrases may be missed.

We normalize case and LaTeX formatting, treating words, numbers, and
mathematical symbols as tokens. We exclude $n$-grams already present in
the original problem. Each retained fragment must contain at least two
alphanumeric tokens, including a non-stopword alphabetic token, and at
least one non-stopword alphanumeric token absent from the original
problem. These filters exclude number-only and single-variable fragments.
The reuse rate $R_m$ for method $m$ is the fraction of responses with a
match, averaged equally across the 25 model--dataset strata.
All methods use the same 9,837 NoOp model--question instances, giving
39,348 saved responses across Base, SL, TL, and Both.
Table~\ref{tab:analysis-summary} uses $n=3$ and reports the relative change
$100(R_m/R_{\mathrm{Base}}-1)$.

\begin{table}[t]
\centering
\small
\setlength{\tabcolsep}{5pt}
\renewcommand{\arraystretch}{1.1}
\begin{tabular}{@{}lrrr@{}}
\toprule
Method & 3-gram & 4-gram & 5-gram \\
\midrule
\multicolumn{4}{l}{\textit{Reuse rate (\%)}} \\
Base & 63.87 & 57.03 & 51.66 \\
SL   & 63.52 & 56.08 & 49.78 \\
TL   & 14.37 & 12.44 & 11.05 \\
Both & 13.52 & 11.27 & 9.87 \\
\midrule
\multicolumn{4}{l}{\textit{Relative change from Base (\%)}} \\
SL   & $-0.6$ & $-1.7$ & $-3.6$ \\
TL   & $-77.5$ & $-78.2$ & $-78.6$ \\
Both & $-78.8$ & $-80.2$ & $-80.9$ \\
\bottomrule
\end{tabular}
\caption{Sensitivity of lexical distractor reuse to $n$-gram length on
the same full NoOp evaluation set. All settings use identical filtering
rules and equally weight the 25 model--dataset strata. Both denotes
SL+TL; lower values indicate less lexical reuse.}
\label{tab:distractor-reuse-sensitivity}
\end{table}

Table~\ref{tab:distractor-reuse-sensitivity} shows the same qualitative
pattern for $n\in\{3,4,5\}$: TL and Both substantially reduce reuse,
whereas SL changes the reuse rate only modestly. Eligible fragments exist
for 9,598, 9,598, and 9,593 model--question instances, respectively;
instances without eligible fragments remain in the denominator and
receive zero. Restricting all three settings to the same 9,588 instances
with eligible fragments at every $n$ preserves this pattern.
Longer $n$-grams can miss short or partially restated conditions, so
these results establish sensitivity of the measured rates and consistency
of the editing effect, rather than the semantic accuracy of a particular
$n$-gram length.

\section{Generation Metrics}
\label{app:saved-trace-efficiency}
This section defines the automatic metrics reported in
Table~\ref{tab:analysis-summary}.
Following prior evaluations of overthinking and reasoning efficiency,
we use generation length, token efficiency, and token-budget exhaustion
\citep{sfwe_ot_kaiser2026}, together with repetition-loop incidence
\citep{sfwe_ot_alimaskina2026}, as quantitative proxies for overthinking.
These measures capture generation cost, correct answers per unit of
generation, budget exhaustion, and lexical repetition, respectively.
They provide indirect evidence about overthinking rather than direct
semantic judgments of whether each reasoning step is unnecessary.

\subsection{Evaluation population and measures}
\paragraph{Population and fixed comparisons.}
We compare Base with SL-only (SL-Attn), TL-only (TL-MLP), and the
fixed TL-neuron/SL-head composition. These use the components of
the combined intervention: all positively attributed SL attention
heads and the top 25\% of positively attributed TL-MLP neurons.
All five models and five datasets are included in both original and
NoOp conditions, with 9,837 paired model--question observations per
condition for each comparison.
The per-model counts are GSM 1,000, AMC 127, Gaokao 385, and pooled
AIME2024--2026 90; Math500 uses the common fitting-excluded evaluation
subset for each model.
Macro-averages weight the five datasets equally within a model and
then the five models equally.

\paragraph{Token cost, efficiency, and cap incidence.}
Let $L_i^B,L_i^E$ be stored API completion-token counts,
$Y_i^B,Y_i^E\in\{0,1\}$ the stored correctness labels, and
$H_i^B,H_i^E$ the stored length-cap indicators for Base and the
edited model. Within each model--dataset cell, we compute mean
token cost, cap incidence, and token efficiency:
\[
 E_0=1000\,\frac{\sum_i Y_i}{\sum_i L_i}.
\]
$E_0$ is correct answers per 1,000 generated tokens
\citep{sfwe_ot_kaiser2026}; it is not an accuracy percentage or an
average of per-instance efficiency ratios.
Tokens, Token Eff., and Cap report $100(M^E/M^B-1)$, where $M$
is the equal-weight mean of the corresponding measure over the
25 model--dataset cells. Thus these columns report relative changes
of macro-averages, not averages of cellwise relative changes.
Token cost uses the complete saved generation and does not constitute
equal physical compute across different model tokenizers.
Correctness labels are unchanged.

\paragraph{Repetition representation.}
The strict-loop detector requires an exact 20-token $n$-gram to
appear at least four times wholly within any sliding span of
1,024 generated-token positions, with overlaps counted
\citep{sfwe_ot_alimaskina2026}.
For each Base--edit pair, we apply the detector to the first
$\min(\widetilde L_i^B,\widetilde L_i^E)$ reencoded tokens of each
response, where $\widetilde L$ denotes its reencoded length.
Loop reports the edit-minus-Base change in loop incidence in
percentage points, averaged equally over the 25 model--dataset cells.
Original generated token IDs are unavailable, so lexical measures
use the corresponding offline tokenizers with no added special
tokens, truncation or padding. Cost always uses stored API counts.
Matched prefixes control the length available to the detector;
these lexical diagnostics do not by themselves establish a change
in semantic overthinking.

\section{Vocabulary Readout}
\label{app:vocabulary-readout}
\paragraph{Projection and ranking.}
We use the saved, layer-specific unit TL and SL directions in their
NoOp-oriented sign for each of the five models. For a model with $L$
transformer blocks indexed from zero, a relative depth $p$ selects
$c=\lfloor p(L-1)\rfloor$ and the three-layer band
$B_p=\{c-1,c,c+1\}$. The primary readout fixes $p=0.75$ for every model;
$p=0.25$ and $0.50$ provide depth-sensitivity readouts using the same rule.
The primary zero-based bands are 19--21 for R1-1.5B and R1-7B,
22--24 for R1-8B, 46--48 for R1-32B, and 25--27 for Qwen3-4B.

For each layer, we compute the full-vocabulary raw projection
$s_\ell=W_U d_\ell$ in float32, using the \texttt{lm\_head} weights of
the four R1 distills and the tied embedding weights of Qwen3-4B.
We sum the three score vectors in ascending layer order in float64 and
divide by three. Tokens are ranked by descending, unrounded mean score,
with ascending token ID breaking exact ties. We apply no sign reversal,
final normalization, output bias, softmax, token filtering, or
deduplication. These scores are projections rather than probabilities.
The readout uses fixed local weight and tokenizer files; it does not
establish their original historical repository revisions.

\paragraph{Display and interpretation.}
Table~\ref{tab:vocab-primary} reports ranks 1--10 at the primary depth;
the first four entries are exactly those in Table~\ref{tab:tltokens}.
Tables~\ref{tab:vocab-quarter} and~\ref{tab:vocab-half} give the same
ranks at 25\% and 50\% depth. Entries run left to right from rank 1 to rank 10,
with superscripts giving token IDs. \texttt{\textvisiblespace}
represents each literal ASCII space, including space-only tokens.
Non-ASCII characters outside the displayed Chinese script use literal
\texttt{\textbackslash uXXXX} Unicode escapes; thus
\texttt{\textbackslash u00a0} is a non-breaking space, not an ordinary
space. Case, fragments, punctuation, and distinct IDs are retained without
semantic selection. The lists are descriptive vocabulary projections;
the behavioral hypotheses are assessed by the separate interventions and
generation analyses in Section~\ref{analysis}.

\begin{table*}[tp]
\centering
\footnotesize
\setlength{\tabcolsep}{4pt}
\renewcommand{\arraystretch}{1.15}
\begin{tabular}{@{}lp{0.415\textwidth}p{0.415\textwidth}@{}}
\toprule
Model / layers & TL: ranks 1--10 & SL: ranks 1--10 \\
\midrule
\shortstack[l]{R1-1.5B\\19--21} &
\texttt{\textvisiblespace{}also}\textsuperscript{1083}, \texttt{\textvisiblespace{}given}\textsuperscript{2661}, \texttt{\textvisiblespace{}mention}\textsuperscript{6286}, \texttt{\textvisiblespace{}additionally}\textsuperscript{36838}, \texttt{also}\textsuperscript{18654}, \texttt{given}\textsuperscript{41968}, \texttt{Additionally}\textsuperscript{49574}, \texttt{Also}\textsuperscript{13394}, \texttt{\begin{CJK}{UTF8}{gbsn}还有一个\end{CJK}}\textsuperscript{107446}, \texttt{\textvisiblespace{}Additionally}\textsuperscript{22406} &
\texttt{OCR}\textsuperscript{93495}, \texttt{ird}\textsuperscript{2603}, \texttt{\textvisiblespace{}\$\#}\textsuperscript{65915}, \texttt{\textvisiblespace{}nevertheless}\textsuperscript{37230}, \texttt{\textvisiblespace{}GX}\textsuperscript{62052}, \texttt{\textvisiblespace{}somehow}\textsuperscript{16925}, \texttt{\textvisiblespace{}souls}\textsuperscript{34726}, \texttt{pll}\textsuperscript{80438}, \texttt{+[}\textsuperscript{69735}, \texttt{\textvisiblespace{}Holocaust}\textsuperscript{43098}\\
\addlinespace[4pt]
\shortstack[l]{R1-7B\\19--21} &
\texttt{\textvisiblespace{}additionally}\textsuperscript{36838}, \texttt{Additionally}\textsuperscript{49574}, \texttt{\textvisiblespace{}unrelated}\textsuperscript{45205}, \texttt{\textvisiblespace{}Additionally}\textsuperscript{22406}, \texttt{\textvisiblespace{}subsidi}\textsuperscript{31519}, \texttt{\textvisiblespace{}furthermore}\textsuperscript{77537}, \texttt{Furthermore}\textsuperscript{56317}, \texttt{Also}\textsuperscript{13394}, \texttt{idata}\textsuperscript{81080}, \texttt{\textvisiblespace{}Also}\textsuperscript{7281} &
\texttt{ieber}\textsuperscript{60241}, \texttt{allet}\textsuperscript{7464}, \texttt{\textvisiblespace{}asn}\textsuperscript{75252}, \texttt{ITU}\textsuperscript{48299}, \texttt{\textvisiblespace{}cauliflower}\textsuperscript{95870}, \texttt{\textvisiblespace{}demi}\textsuperscript{92028}, \texttt{\textvisiblespace{}tant}\textsuperscript{36522}, \texttt{\textvisiblespace{}cru}\textsuperscript{22218}, \texttt{\textvisiblespace{}disconnected}\textsuperscript{39086}, \texttt{\textvisiblespace{}Escape}\textsuperscript{45643}\\
\addlinespace[4pt]
\shortstack[l]{R1-8B\\22--24} &
\texttt{given}\textsuperscript{43068}, \texttt{\textvisiblespace{}given}\textsuperscript{2728}, \texttt{\textvisiblespace{}Given}\textsuperscript{16644}, \texttt{Given}\textsuperscript{22818}, \texttt{Also}\textsuperscript{13699}, \texttt{\textvisiblespace{}also}\textsuperscript{1101}, \texttt{\textvisiblespace{}ALSO}\textsuperscript{50921}, \texttt{\textvisiblespace{}Also}\textsuperscript{7429}, \texttt{\_given}\textsuperscript{58753}, \texttt{\textvisiblespace{}additionally}\textsuperscript{37938} &
\texttt{\textvisiblespace{}Cor}\textsuperscript{4563}, \texttt{idebar}\textsuperscript{12972}, \texttt{anche}\textsuperscript{29097}, \texttt{\textvisiblespace{}Devin}\textsuperscript{81092}, \texttt{\textvisiblespace{}tables}\textsuperscript{12920}, \texttt{\textvisiblespace{}again}\textsuperscript{1578}, \texttt{\textvisiblespace{}}\textsuperscript{220}, \texttt{\textvisiblespace{}tang}\textsuperscript{22636}, \texttt{\textvisiblespace{}Gio}\textsuperscript{67765}, \texttt{955}\textsuperscript{25875}\\
\addlinespace[4pt]
\shortstack[l]{R1-32B\\46--48} &
\texttt{\begin{CJK}{UTF8}{gbsn}另有\end{CJK}}\textsuperscript{108427}, \texttt{\begin{CJK}{UTF8}{gbsn}无关\end{CJK}}\textsuperscript{107173}, \texttt{\begin{CJK}{UTF8}{gbsn}另\end{CJK}}\textsuperscript{99628}, \texttt{\begin{CJK}{UTF8}{gbsn}有个\end{CJK}}\textsuperscript{104627}, \texttt{\textvisiblespace{}Additionally}\textsuperscript{22406}, \texttt{\begin{CJK}{UTF8}{gbsn}另外\end{CJK}}\textsuperscript{101948}, \texttt{\begin{CJK}{UTF8}{gbsn}有一个\end{CJK}}\textsuperscript{104133}, \texttt{\textvisiblespace{}additionally}\textsuperscript{36838}, \texttt{\begin{CJK}{UTF8}{gbsn}辅\end{CJK}}\textsuperscript{100519}, \texttt{\begin{CJK}{UTF8}{gbsn}同时\end{CJK}}\textsuperscript{91572} &
\texttt{arend}\textsuperscript{52845}, \texttt{/swagger}\textsuperscript{80930}, \texttt{\textvisiblespace{}Berg}\textsuperscript{30682}, \texttt{\begin{CJK}{UTF8}{gbsn}无论如何\end{CJK}}\textsuperscript{113232}, \texttt{\textvisiblespace{}Yang}\textsuperscript{24474}, \texttt{\begin{CJK}{UTF8}{gbsn}网站地图\end{CJK}}\textsuperscript{110850}, \texttt{\textvisiblespace{}sod}\textsuperscript{23952}, \texttt{Bid}\textsuperscript{65452}, \texttt{IFIED}\textsuperscript{27320}, \texttt{.LoggerFactory}\textsuperscript{35828}\\
\addlinespace[4pt]
\shortstack[l]{Qwen3-4B\\25--27} &
\texttt{\begin{CJK}{UTF8}{gbsn}还有一个\end{CJK}}\textsuperscript{107446}, \texttt{\textvisiblespace{}given}\textsuperscript{2661}, \texttt{\begin{CJK}{UTF8}{gbsn}提及\end{CJK}}\textsuperscript{109361}, \texttt{\begin{CJK}{UTF8}{gbsn}提到\end{CJK}}\textsuperscript{104496}, \texttt{\begin{CJK}{UTF8}{gbsn}另有\end{CJK}}\textsuperscript{108427}, \texttt{\begin{CJK}{UTF8}{gbsn}有个\end{CJK}}\textsuperscript{104627}, \texttt{\textvisiblespace{}mention}\textsuperscript{6286}, \texttt{\textvisiblespace{}also}\textsuperscript{1083}, \texttt{\begin{CJK}{UTF8}{gbsn}另外\end{CJK}}\textsuperscript{101948}, \texttt{\begin{CJK}{UTF8}{gbsn}有一个\end{CJK}}\textsuperscript{104133} &
\texttt{icorn}\textsuperscript{50356}, \texttt{\begin{CJK}{UTF8}{gbsn}特别声明\end{CJK}}\textsuperscript{116651}, \texttt{eyJ}\textsuperscript{84609}, \texttt{\begin{CJK}{UTF8}{gbsn}某种\end{CJK}}\textsuperscript{102988}, \texttt{\begin{CJK}{UTF8}{gbsn}有价值的\end{CJK}}\textsuperscript{111186}, \texttt{ething}\textsuperscript{2298}, \texttt{\textvisiblespace{}Exercises}\textsuperscript{90454}, \texttt{oplay}\textsuperscript{36265}, \texttt{\begin{CJK}{UTF8}{gbsn}的重点\end{CJK}}\textsuperscript{108224}, \texttt{\begin{CJK}{UTF8}{gbsn}我真的\end{CJK}}\textsuperscript{110146}\\
\bottomrule
\end{tabular}
\caption{Top-10 tokens for the 75\%-depth primary readout, ranked by the
unrounded mean raw projection over the listed zero-based layer band.
Superscripts are token IDs, not scores. Spaces and Unicode escapes follow
the display convention in Appendix~\ref{app:vocabulary-readout}; no
filtering or deduplication is applied.}
\label{tab:vocab-primary}
\end{table*}

\begin{table*}[tp]
\centering
\footnotesize
\setlength{\tabcolsep}{4pt}
\renewcommand{\arraystretch}{1.15}
\begin{tabular}{@{}lp{0.415\textwidth}p{0.415\textwidth}@{}}
\toprule
Model / layers & TL: ranks 1--10 & SL: ranks 1--10 \\
\midrule
\shortstack[l]{R1-1.5B\\5--7} &
\texttt{\textvisiblespace{}Worlds}\textsuperscript{51847}, \texttt{\begin{CJK}{UTF8}{gbsn}背面\end{CJK}}\textsuperscript{116522}, \texttt{\textvisiblespace{}now}\textsuperscript{1431}, \texttt{\textbackslash{}u2243\allowbreak{}}\textsuperscript{148158}, \texttt{\textvisiblespace{}next}\textsuperscript{1790}, \texttt{-target}\textsuperscript{18489}, \texttt{\textasciitilde{}}\textsuperscript{93}, \texttt{\textvisiblespace{}?:}\textsuperscript{27257}, \texttt{\textvisiblespace{}DAG}\textsuperscript{79987}, \texttt{now}\textsuperscript{3328} &
\texttt{\begin{CJK}{UTF8}{gbsn}狼\end{CJK}}\textsuperscript{104522}, \texttt{iah}\textsuperscript{19310}, \texttt{\textvisiblespace{}role}\textsuperscript{3476}, \texttt{ernet}\textsuperscript{13845}, \texttt{\textvisiblespace{}std}\textsuperscript{1460}, \texttt{gram}\textsuperscript{1520}, \texttt{yk}\textsuperscript{72540}, \texttt{-query}\textsuperscript{65489}, \texttt{\textvisiblespace{}TXT}\textsuperscript{72429}, \texttt{\textvisiblespace{}cru}\textsuperscript{22218}\\
\addlinespace[4pt]
\shortstack[l]{R1-7B\\5--7} &
\texttt{\textvisiblespace{}justification}\textsuperscript{41435}, \texttt{\textvisiblespace{}moot}\textsuperscript{97783}, \texttt{inosaur}\textsuperscript{88147}, \texttt{\textvisiblespace{}MSP}\textsuperscript{68373}, \texttt{\textvisiblespace{}SPD}\textsuperscript{62724}, \texttt{inode}\textsuperscript{52239}, \texttt{ARGV}\textsuperscript{60657}, \texttt{\textvisiblespace{}dipped}\textsuperscript{72207}, \texttt{\textvisiblespace{}digs}\textsuperscript{88957}, \texttt{\textvisiblespace{}justified}\textsuperscript{34416} &
\texttt{\textvisiblespace{}tink}\textsuperscript{90584}, \texttt{)?.}\textsuperscript{43507}, \texttt{\textvisiblespace{}Reach}\textsuperscript{49842}, \texttt{\textvisiblespace{}dummy}\textsuperscript{17292}, \texttt{\begin{CJK}{UTF8}{gbsn}判定\end{CJK}}\textsuperscript{112983}, \texttt{rose}\textsuperscript{24847}, \texttt{\textvisiblespace{}vice}\textsuperscript{16774}, \texttt{alto}\textsuperscript{75677}, \texttt{\begin{CJK}{UTF8}{gbsn}问候\end{CJK}}\textsuperscript{113940}, \texttt{Cod}\textsuperscript{63691}\\
\addlinespace[4pt]
\shortstack[l]{R1-8B\\6--8} &
\texttt{\textbackslash{}u0430\allowbreak{}\textbackslash{}u0441\allowbreak{}}\textsuperscript{18437}, \texttt{\textbackslash{}u043b\allowbreak{}\textbackslash{}u0438\allowbreak{}\textbackslash{}u043d\allowbreak{}\textbackslash{}u0430\allowbreak{}}\textsuperscript{113891}, \texttt{chied}\textsuperscript{32822}, \texttt{b}\textsuperscript{65}, \texttt{bio}\textsuperscript{38423}, \texttt{arget}\textsuperscript{1307}, \texttt{uja}\textsuperscript{89808}, \texttt{UDO}\textsuperscript{87738}, \texttt{\textvisiblespace{}eleg}\textsuperscript{18190}, \texttt{uyo}\textsuperscript{87788} &
\texttt{efa}\textsuperscript{76731}, \texttt{\textvisiblespace{}s\textbackslash{}u00e1\allowbreak{}m}\textsuperscript{112472}, \texttt{\textvisiblespace{}mastur}\textsuperscript{79202}, \texttt{\textvisiblespace{}\textbackslash{}u062c\allowbreak{}\textbackslash{}u0645\allowbreak{}\textbackslash{}u0647\allowbreak{}\textbackslash{}u0648\allowbreak{}\textbackslash{}u0631\allowbreak{}}\textsuperscript{122078}, \texttt{uteur}\textsuperscript{40095}, \texttt{olumn}\textsuperscript{1309}, \texttt{ynam}\textsuperscript{11694}, \texttt{u\textbackslash{}u015f\allowbreak{}}\textsuperscript{96632}, \texttt{\textbackslash{}u00e1\allowbreak{}genes}\textsuperscript{87108}, \texttt{\textbackslash{}u00fd\allowbreak{}t}\textsuperscript{97700}\\
\addlinespace[4pt]
\shortstack[l]{R1-32B\\14--16} &
\texttt{\begin{CJK}{UTF8}{gbsn}白马\end{CJK}}\textsuperscript{115866}, \texttt{\begin{CJK}{UTF8}{gbsn}一行\end{CJK}}\textsuperscript{105643}, \texttt{\begin{CJK}{UTF8}{gbsn}赎\end{CJK}}\textsuperscript{103815}, \texttt{\begin{CJK}{UTF8}{gbsn}省内\end{CJK}}\textsuperscript{109491}, \texttt{\begin{CJK}{UTF8}{gbsn}灾\end{CJK}}\textsuperscript{99969}, \texttt{\textvisiblespace{}Ellison}\textsuperscript{81647}, \texttt{\begin{CJK}{UTF8}{gbsn}泷\end{CJK}}\textsuperscript{119948}, \texttt{ROS}\textsuperscript{33439}, \texttt{\begin{CJK}{UTF8}{gbsn}马丁\end{CJK}}\textsuperscript{116916}, \texttt{\begin{CJK}{UTF8}{gbsn}声音\end{CJK}}\textsuperscript{102274} &
\texttt{\textvisiblespace{}K\textbackslash{}u0131\allowbreak{}}\textsuperscript{143804}, \texttt{prd}\textsuperscript{86222}, \texttt{\begin{CJK}{UTF8}{gbsn}已经有\end{CJK}}\textsuperscript{108489}, \texttt{\textbackslash{}u1d40\allowbreak{}}\textsuperscript{147202}, \texttt{\textbackslash{}u062a\allowbreak{}\textbackslash{}u0648\allowbreak{}\textbackslash{}u062c\allowbreak{}}\textsuperscript{137230}, \texttt{GenerationStrategy}\textsuperscript{76545}, \texttt{t\textbackslash{}u00f6\allowbreak{}r}\textsuperscript{130457}, \texttt{\begin{CJK}{UTF8}{gbsn}与否\end{CJK}}\textsuperscript{114852}, \texttt{\textvisiblespace{}Levi}\textsuperscript{55041}, \texttt{\begin{CJK}{UTF8}{gbsn}在那里\end{CJK}}\textsuperscript{106504}\\
\addlinespace[4pt]
\shortstack[l]{Qwen3-4B\\7--9} &
\texttt{pulse}\textsuperscript{78207}, \texttt{kap}\textsuperscript{38834}, \texttt{\textvisiblespace{}imp}\textsuperscript{3163}, \texttt{choice}\textsuperscript{11746}, \texttt{ataire}\textsuperscript{65649}, \texttt{\_pulse}\textsuperscript{83645}, \texttt{Ton}\textsuperscript{69451}, \texttt{ULSE}\textsuperscript{83389}, \texttt{\begin{CJK}{UTF8}{gbsn}其中之一\end{CJK}}\textsuperscript{117548}, \texttt{Nit}\textsuperscript{97447} &
\texttt{river}\textsuperscript{5469}, \texttt{\begin{CJK}{UTF8}{gbsn}相关政策\end{CJK}}\textsuperscript{116009}, \texttt{\textvisiblespace{}meant}\textsuperscript{8791}, \texttt{\textvisiblespace{}swell}\textsuperscript{77253}, \texttt{(/\textasciicircum{}}\textsuperscript{50048}, \texttt{\begin{CJK}{UTF8}{gbsn}朝廷\end{CJK}}\textsuperscript{111827}, \texttt{vil}\textsuperscript{45252}, \texttt{retrieve}\textsuperscript{55527}, \texttt{\begin{CJK}{UTF8}{gbsn}闪电\end{CJK}}\textsuperscript{112182}, \texttt{\begin{CJK}{UTF8}{gbsn}突然\end{CJK}}\textsuperscript{103961}\\
\bottomrule
\end{tabular}
\caption{Top-10 tokens for the 25\%-depth sensitivity readout, ranked by the
unrounded mean raw projection over the listed zero-based layer band.
Superscripts are token IDs, not scores. Spaces and Unicode escapes follow
the display convention in Appendix~\ref{app:vocabulary-readout}; no
filtering or deduplication is applied.}
\label{tab:vocab-quarter}
\end{table*}

\begin{table*}[tp]
\centering
\footnotesize
\setlength{\tabcolsep}{4pt}
\renewcommand{\arraystretch}{1.15}
\begin{tabular}{@{}lp{0.415\textwidth}p{0.415\textwidth}@{}}
\toprule
Model / layers & TL: ranks 1--10 & SL: ranks 1--10 \\
\midrule
\shortstack[l]{R1-1.5B\\12--14} &
\texttt{\textvisiblespace{}then}\textsuperscript{1221}, \texttt{\textvisiblespace{}secretly}\textsuperscript{41739}, \texttt{oret}\textsuperscript{18174}, \texttt{then}\textsuperscript{3391}, \texttt{adius}\textsuperscript{4037}, \texttt{\textvisiblespace{}remotely}\textsuperscript{38429}, \texttt{\textvisiblespace{}Woods}\textsuperscript{34748}, \texttt{QUOTE}\textsuperscript{75543}, \texttt{\textvisiblespace{}ordin}\textsuperscript{48642}, \texttt{ycle}\textsuperscript{5449} &
\texttt{\textvisiblespace{}again}\textsuperscript{1549}, \texttt{.}\textsuperscript{13}, \texttt{\textbackslash{}u00a0\allowbreak{}}\textsuperscript{4102}, \texttt{Diamond}\textsuperscript{76510}, \texttt{\textvisiblespace{}\textbackslash{}u2026\allowbreak{}.}\textsuperscript{65521}, \texttt{\textvisiblespace{}Sure}\textsuperscript{22555}, \texttt{\textvisiblespace{}en}\textsuperscript{662}, \texttt{\textvisiblespace{}somehow}\textsuperscript{16925}, \texttt{\textvisiblespace{}\textbackslash{}u201c\allowbreak{}\textbackslash{}u2026\allowbreak{}}\textsuperscript{91013}, \texttt{poi}\textsuperscript{82263}\\
\addlinespace[4pt]
\shortstack[l]{R1-7B\\12--14} &
\texttt{\textvisiblespace{}unrelated}\textsuperscript{45205}, \texttt{.equals}\textsuperscript{6488}, \texttt{\begin{CJK}{UTF8}{gbsn}也为\end{CJK}}\textsuperscript{109001}, \texttt{\begin{CJK}{UTF8}{gbsn}耗\end{CJK}}\textsuperscript{100293}, \texttt{Rooms}\textsuperscript{58259}, \texttt{\textvisiblespace{}Permit}\textsuperscript{80639}, \texttt{\begin{CJK}{UTF8}{gbsn}这张\end{CJK}}\textsuperscript{108893}, \texttt{anye}\textsuperscript{48283}, \texttt{eded}\textsuperscript{29935}, \texttt{\begin{CJK}{UTF8}{gbsn}才有\end{CJK}}\textsuperscript{106119} &
\texttt{onas}\textsuperscript{55301}, \texttt{\textvisiblespace{}iceberg}\textsuperscript{88035}, \texttt{\textvisiblespace{}Zusammen}\textsuperscript{76604}, \texttt{\textvisiblespace{}Validates}\textsuperscript{97843}, \texttt{\textvisiblespace{}competit}\textsuperscript{51204}, \texttt{\textvisiblespace{}rated}\textsuperscript{21628}, \texttt{?:}\textsuperscript{4820}, \texttt{\textvisiblespace{}uplift}\textsuperscript{96447}, \texttt{\textvisiblespace{}\begin{CJK}{UTF8}{gbsn}默认\end{CJK}}\textsuperscript{80546}, \texttt{pla}\textsuperscript{56538}\\
\addlinespace[4pt]
\shortstack[l]{R1-8B\\14--16} &
\texttt{\textvisiblespace{}ALSO}\textsuperscript{50921}, \texttt{\textvisiblespace{}reference}\textsuperscript{5905}, \texttt{\begin{CJK}{UTF8}{gbsn}还有\end{CJK}}\textsuperscript{114002}, \texttt{\textvisiblespace{}also}\textsuperscript{1101}, \texttt{utow}\textsuperscript{9813}, \texttt{anda}\textsuperscript{10018}, \texttt{\textvisiblespace{}\textbackslash{}u0442\allowbreak{}\textbackslash{}u0430\allowbreak{}\textbackslash{}u043a\allowbreak{}\textbackslash{}u0436\allowbreak{}\textbackslash{}u0435\allowbreak{}}\textsuperscript{102277}, \texttt{\textvisiblespace{}meanwhile}\textsuperscript{37318}, \texttt{Also}\textsuperscript{13699}, \texttt{\textvisiblespace{}PLUS}\textsuperscript{59553} &
\texttt{yx}\textsuperscript{52926}, \texttt{955}\textsuperscript{25875}, \texttt{\textbackslash{}uad81\allowbreak{}}\textsuperscript{121321}, \texttt{Connecting}\textsuperscript{64024}, \texttt{\textvisiblespace{}Devin}\textsuperscript{81092}, \texttt{\textvisiblespace{}Rig}\textsuperscript{52632}, \texttt{gis}\textsuperscript{52198}, \texttt{\textvisiblespace{}pending}\textsuperscript{15639}, \texttt{\textbackslash{}u0435\allowbreak{}\textbackslash{}u043a\allowbreak{}\textbackslash{}u0442\allowbreak{}}\textsuperscript{30656}, \texttt{\textvisiblespace{}pole}\textsuperscript{26078}\\
\addlinespace[4pt]
\shortstack[l]{R1-32B\\30--32} &
\texttt{\begin{CJK}{UTF8}{gbsn}多彩\end{CJK}}\textsuperscript{103585}, \texttt{AndUpdate}\textsuperscript{56365}, \texttt{\_STAR}\textsuperscript{70786}, \texttt{\begin{CJK}{UTF8}{gbsn}骞\end{CJK}}\textsuperscript{121318}, \texttt{\begin{CJK}{UTF8}{gbsn}醮\end{CJK}}\textsuperscript{122010}, \texttt{\textvisiblespace{}\_|}\textsuperscript{85137}, \texttt{\begin{CJK}{UTF8}{gbsn}琐\end{CJK}}\textsuperscript{103816}, \texttt{\begin{CJK}{UTF8}{gbsn}煊\end{CJK}}\textsuperscript{121291}, \texttt{\begin{CJK}{UTF8}{gbsn}学到\end{CJK}}\textsuperscript{109741}, \texttt{\begin{CJK}{UTF8}{gbsn}北路\end{CJK}}\textsuperscript{115694} &
\texttt{\textvisiblespace{}Bust}\textsuperscript{70333}, \texttt{ardash}\textsuperscript{55688}, \texttt{\textbackslash{}Migration}\textsuperscript{31613}, \texttt{\_https}\textsuperscript{26817}, \texttt{\textvisiblespace{}Laf}\textsuperscript{60844}, \texttt{ariat}\textsuperscript{40879}, \texttt{\textvisiblespace{}@[}\textsuperscript{95800}, \texttt{\textbackslash{}u25be\allowbreak{}}\textsuperscript{144958}, \texttt{tid}\textsuperscript{24449}, \texttt{rances}\textsuperscript{94497}\\
\addlinespace[4pt]
\shortstack[l]{Qwen3-4B\\16--18} &
\texttt{\textvisiblespace{}pl}\textsuperscript{625}, \texttt{\begin{CJK}{UTF8}{gbsn}染\end{CJK}}\textsuperscript{99702}, \texttt{-linux}\textsuperscript{46375}, \texttt{\begin{CJK}{UTF8}{gbsn}曝\end{CJK}}\textsuperscript{102378}, \texttt{mdir}\textsuperscript{69695}, \texttt{pl}\textsuperscript{500}, \texttt{\textvisiblespace{}im}\textsuperscript{732}, \texttt{\textvisiblespace{}est}\textsuperscript{1788}, \texttt{\begin{CJK}{UTF8}{gbsn}东部\end{CJK}}\textsuperscript{106565}, \texttt{\begin{CJK}{UTF8}{gbsn}曼\end{CJK}}\textsuperscript{101267} &
\texttt{\begin{CJK}{UTF8}{gbsn}万辆\end{CJK}}\textsuperscript{107883}, \texttt{uable}\textsuperscript{13128}, \texttt{\textvisiblespace{}ba\textbackslash{}u015f\allowbreak{}ka}\textsuperscript{129149}, \texttt{.providers}\textsuperscript{87179}, \texttt{.Companion}\textsuperscript{94483}, \texttt{\textvisiblespace{}envi}\textsuperscript{49264}, \texttt{\begin{CJK}{UTF8}{gbsn}争创\end{CJK}}\textsuperscript{118096}, \texttt{sounds}\textsuperscript{70531}, \texttt{.getBoolean}\textsuperscript{44711}, \texttt{youtu}\textsuperscript{54860}\\
\bottomrule
\end{tabular}
\caption{Top-10 tokens for the 50\%-depth sensitivity readout, ranked by the
unrounded mean raw projection over the listed zero-based layer band.
Superscripts are token IDs, not scores. Spaces and Unicode escapes follow
the display convention in Appendix~\ref{app:vocabulary-readout}; no
filtering or deduplication is applied.}
\label{tab:vocab-half}
\end{table*}

\end{document}